\documentclass{article}

\usepackage{arxiv}

\usepackage[utf8]{inputenc} % allow utf-8 input
\usepackage[T1]{fontenc}    % use 8-bit T1 fonts
\usepackage{hyperref}       % hyperlinks
\usepackage{url}            % simple URL typesetting
\usepackage{booktabs}       % professional-quality tables
\usepackage{amsfonts}       % blackboard math symbols
\usepackage{nicefrac}       % compact symbols for 1/2, etc.
\usepackage{microtype}      % microtypography

\usepackage{graphicx}
\usepackage{float}

\usepackage{amsmath}
\usepackage{amssymb}

\usepackage{graphicx}
\usepackage{color}
\newtheorem{proc}{Procedure}
\newtheorem{proposition}{Proposition}
\newtheorem{theorem}{Theorem}

\newtheorem{remark}{Remark}
\newtheorem{corollary}{Corollary}
\newcommand*{\qed}{\hfill\ensuremath{\square}}

\title{Topological signature for periodic  motion recognition}

\author{
  Javier Lamar-Leon, Edel Garcia-Reyes\\  CENATAV, La Havana, Cuba\\         email: \{jlamar, egarcia\}@cenatav.co.cu
\And
Rocio Gonzalez-Diaz\\ School of Computer Engineering,  University of Seville, Spain\\  http:// personal.us.es/rogodi  \\      email: rogodi@us.es }

\begin{document}
\maketitle

\begin{abstract}
In this paper, we present an algorithm that computes the topological signature for a given periodic motion sequence. Such signature consists of a vector obtained by persistent homology which captures the topological and geometric changes of the object that models the motion.
Two topological signatures are compared simply by the angle between the corresponding vectors.
With respect to gait recognition, we have tested our method using only the lowest fourth part of the body's silhouette.
In this way, the impact of variations in the upper part of the body, which are  very frequent in real scenarios, decreases considerably.
We have also tested our method using other periodic motions such as running or jumping.
Finally, we formally prove that our method is robust to small perturbations in the input data and does not depend on the number of periods contained in the periodic motion sequence.
\end{abstract}

% keywords can be removed

\section{Introduction}
Person recognition at distance, without the subject cooperation, is an important task in video surveillance. 
Nevertheless, very few biometric techniques can be used in such scenario. 
Gait recognition is a technique with special potential under these circumstances  since  features can be extracted from any viewpoint and at bigger distances than other biometric approaches. 
Currently, there are good results in the state of the art for persons walking under natural conditions (without carrying a bag or wearing a coat). See, for example,  \cite{bb65346,lee2014time,lamar2012human}. 
However, it is  common for people to walk  carrying things that change their natural gait.
The accuracy in gait recognition for persons carrying bag or using coat  can be consulted, for example, in \cite{bb65346}
for the CASIA-B gait dataset\footnote{http://www.cbsr.ia.ac.cn/GaitDatasetB-silh.zip}.
Moreover, people usually perform movements with the upper body part unrelated to the natural dynamic of the gait.

Up to now, the most successful approaches in gait recognition  use silhouettes to get the features.
 Among the silhouette-based techniques, the best results have been obtained from the methods based in Gait Energy Images (GEI) \cite{zhang2013score,bb65346,lee2014time,rida2016gait,wu2017comprehensive}.  
Generally, these strategies are affected by a small number of silhouettes (one gait cycle or less). 
Moreover, the temporal order in which silhouettes appear is not captured in those representations, losing the relative relations of the movements in time. 
Besides, the features extracted by those methods are highly correlated with errors in the segmentation of the silhouettes \cite{bb68418} and these errors frequently appear in the existing algorithms for background segmentation. This implies that GEI  methods are influenced by the shape of the silhouettes instead of the relative positions among the parts of the body while walking. 

\begin{figure}[ht!]
\centering
\includegraphics[width=2 in]{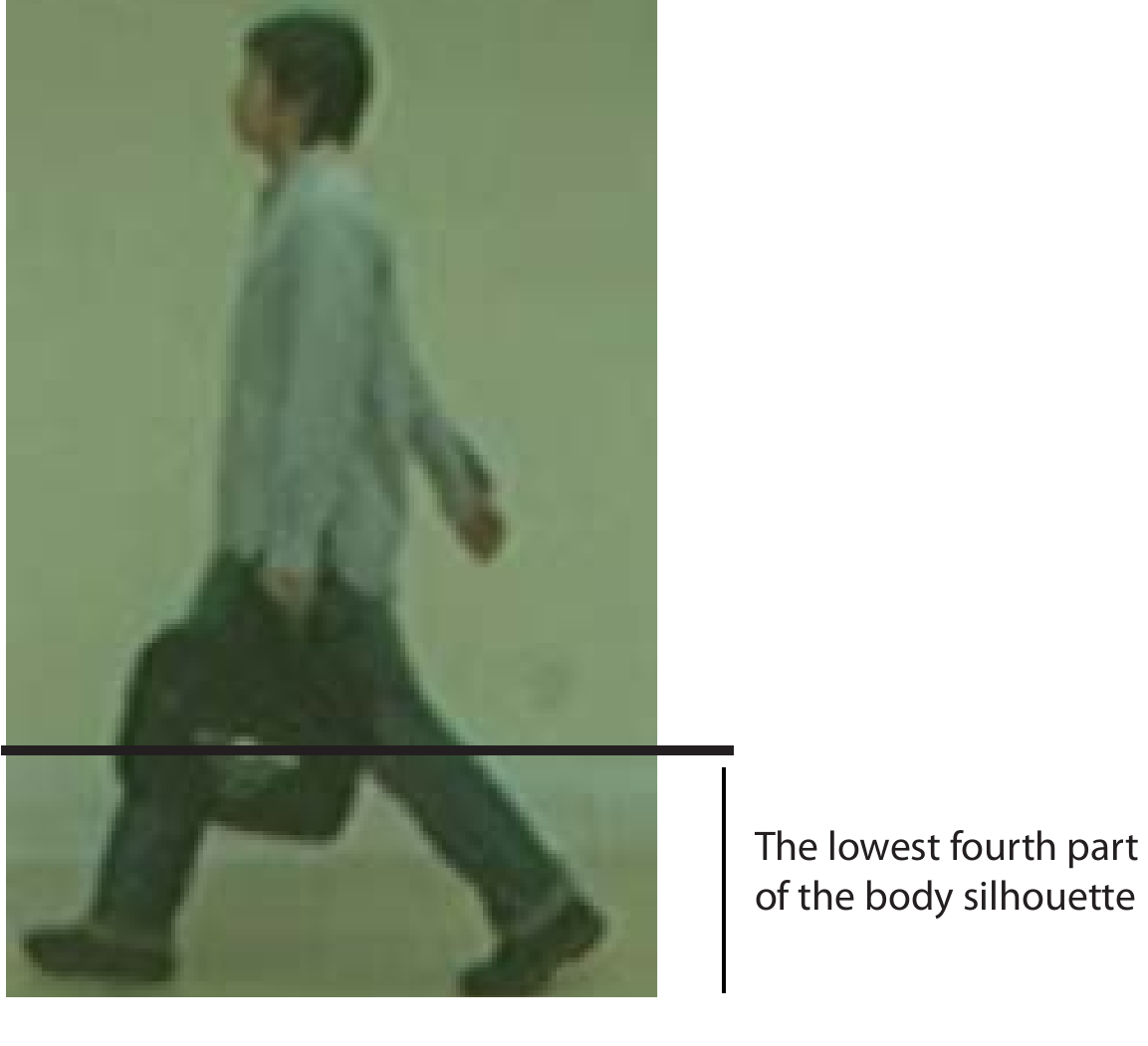} %
\caption{Lowest four part of the body occluded by a bag.}
\label{afect_bag}
\end{figure}

In our  previous conference papers \cite{lamar2012human,leon2013gait,ciarp2014,icpr2014}, we concentrated our effort in overcoming most of the difficulties explained above.
In these works, the gaits were modeled by a  persistent-homology-based representation, called {\it topological signature for the gait sequence}, and used for human identification  \cite{lamar2012human}, gender classification
 \cite{leon2013gait},  carried object detection  \cite{ciarp2014} and monitoring human activities at distance  \cite{icpr2014}.
Later, in \cite{icpr2016}, we computed our topological signature using only  the lower part of the body (see Figure \ref{afect_bag}), avoiding many of the effects arising from the variability in the upper body part (related, for example,  to hand  gestures while talking on cell). 
This selection is  endorsed by the result given in \cite{bashir2010gait}, which shows that this part of the body provides most of the necessary information for classification.
The topological signature defined in our previous papers has been also used in \cite{yonghzhen} for recognizing 3D face expression and  \cite{MJ} for differentiating  forehand
and backhand strokes  performed by  a tennis player.

 Other papers  using persistent homology for action recognition are the followings.
 In \cite{RELATED2016},  topological features of the attractor of the dynamical
system are used for modelling human actions and  a nearest neighbor classifier is trained with the  persistence-based features. In  \cite{related2016-1}, 
 point clouds describing the oscillatory patterns of body joints from the principal components of their time series using Taken's delay embedding are computed. In \cite{related2015}, 
 a novel framework, based on persistent cohomology can automatically detect,
parameterize and interpolate periodic motion patterns obtained from a motion  sequence.
 In none of those papers, theoretical results on the correct behaviour of the designed methods are provided.

 In this paper, in Section \ref{preliminaries}, we first recall
 the main background needed to understand it. 
 We generalize the procedure given in our previous papers to obtain the  topological signature for a periodic  motion in Section \ref{simpcompsec}. 
 The input of the  procedure is a sequence of  silhouettes obtained from a video. A simplicial complex $\partial K(I)$
which represents the periodic motion is then constructed in Subsection \ref{simplicial}.
Sixteen  persistence barcodes are then computed  considering the distance to eight fixed planes: $2$ horizontal, $2$ vertical, $2$ oblique and $2$ depth planes, completely capturing, this way, 
 the motion in the sequence (see Subsection \ref{filtration}).
 More concretely, for each plane $\pi$, we compute two persistence barcodes (graphical tools encoding the persistent homology information).  One persistence barcode  detects the variation of connected components and the other one detects the variation of tunnels, when we go through  $\partial K(I)$ in a direction  perpendicular to the plane $\pi$.  
Putting together all this information, we construct a vector called {\it topological signature} for each sequence in Subsection \ref{signat}. We
compare two topological signatures by  the angle between  the vectors forming  the signatures. 
 As an original contribution of this paper, we  theoretically study  the stability of the topological signature in Section \ref{section:stability}. We formally prove, in terms of probabilities, that small perturbations in the input body silhouettes provoke small perturbations in the resulting topological signature. We also prove that the direction of each of the vectors that make up the topological signature  for a  sequence remains the same independently on the number of  periods the sequence contains. Since we compare two topological signatures by the angle between the corresponding vectors, then the previous assertion implies that the topological signature is independent on the number of  periods the sequence contains. 
Experimental results are showed in Section \ref{exper}. Conclusions are   given in Section \ref{conclusions}.

\section{Preliminaries}\label{preliminaries}

Let us now introduce the  definitions and concepts used throughout the paper. The main one is the concept of persistent homology, which is an algebraic tool for measuring topological features of shapes and functions. It is built on top of homology, which is a topological invariant that captures the amount of connected components, tunnels, cavities and higher-dimensional counterparts of a shape. Small size features in persistent homology are often categorized as noise, while large size features describe topological properties of shapes.   For a more detailed introduction on the theoretical concepts introduced in this section see, for example, \cite{compTopologyBook,ghrist,chazal}.

%%%%%%%%%%%%%%%%%%%%%%%%%%%%%%%%%%%%%%%%%%%%%%%%%%

A {\it 3D binary  image}
is a pair  $I=(\mathbb{Z}^3,B)$, where $B$ (called the {\itshape foreground}) is a finite set of points of $\mathbb{Z}^3$ and  $B^c=\mathbb{Z}^3 \backslash B$ is the {\itshape background}.
The {\it cubical complex} $Q(I)$ associated to  $I$ is a  combinatorial structure constituted by a
set of unit cubes with square faces parallel to the coordinate planes and vertices in $\mathbb{Z}^3$.
More concretely, the set of vertices
$V$ of any cube $c\in Q(I)$ satisfies that
$V = \{(i, j, k), (i+1, j, k), (i, j +1, k), (i, j, k+1), (i+1, j+1, k), (i+1, j, k+1),(i, j+1, k+1), (i+1, j+1, k+1)\}$ for some $(i, j, k) \in \mathbb{Z}^3$, and  $V \subseteq B$.
The $0$-{\it faces} of  $c$ are its $8$ corners (vertices), its $1$-faces are its $12$ edges, its $2$-faces are its $6$ squares and, finally, its $3$-face is the cube $c$ itself. 
 
%%%%%%%%%%%%%%%%%%%%%%%5%%%%%%%

A {\it $p$-simplex} $\sigma$ in $\mathbb{R}^n$ is the set of $p+1$ affinely independent points in $\mathbb{R}^n$. Observe that always $p\leq n$. 
A set $\mu$ in  $\mathbb{R}^n$ is a {\it face} of $\sigma$ if  $\mu\subseteq \sigma$.
The simplices considered in this paper are $0$-simplices (representing vertices), $1$-simplices (representing edges) and $2$-simplices (representing triangles), all of them embedded in $\mathbb{R}^3$. 
The formal definition of  a simplicial complex $K$ 
 is as follows \cite[p. 7]{mukres}: A {\it simplicial complex} $K$ is a collection of simplices  such that: (1) every face of a simplex of $K$ is in $K$; and (2) the intersection of any two simplices of $K$ is a face of each of them. A simplicial complex is {\it finite} if it has a finite number of simplices.

%%%%%%%%%%%%%%%%%%%%%%%%%%%%%%%%%%%%%%%%%%%%%%%%

Let $K$ be a simplicial complex. A $p$-chain on $K$ is a formal sum of $p$-simplices of $K$. The group of $p$-chains is denoted by $C_p(K)$. 
The $p$-{\it boundary operator} 
 $\partial_p:C_p(K)\to C_{p-1}(K)$ 
 is a  homomorphism  such that for each $p$-simplex $\sigma$ of $K$, $\partial_p(\sigma)$ is the sum of its $(p-1)$-faces. For example, if $\sigma$ is a triangle, $\partial_2(\sigma)$ is the sum of its edges. The kernel of $\partial_p$ is called the group of {\it $p$-cycles} in $C_p(K)$ and the image of $\partial_{p+1}$ is called the group of {\it $p$-boundaries} in $C_{p}(K)$. The {\it $p$-homology} $H_p(K)$ of $K$ is the quotient group of $p$-cycles relative to $p$-boundaries (see \cite[Chapter 5]{mukres}). 
The $0$-homology classes of $K$ (i.e. the classes in $H_0(K)$) represent the connected components of $K$, the $1$-homology classes  its tunnels  and the $2$-homology classes its cavities.

%%%%%%%%%%%%%%%%%%%%%%%%%%%%%%%%%%

A {\it filtration} $F$ of a simplicial complex $K$ is an ordering of the simplices of $K$
 dictated by a {\it filter function} $f:K \rightarrow \mathbb{R}$, satisfying that if a simplex $\sigma$  is a face of another simplex $\sigma'$ in $K$ then $f(\sigma)\leq f(\sigma')$ and   $\sigma$ appears before  $\sigma'$ in the  filtration. The associated {\it filtered simplicial complex} is the sequence:
 $$\emptyset\subset K_{i_1}\subset \cdots \subset K_{i_{\ell}}=K$$
 where $i_1<\cdots<i_{\ell}$ and $K_{i_j}=f^{-1}(-\infty,i_j]$ for $1\leq j\leq \ell$.
 
%%%%%%%%%%%%%%%%%%%%%%%%%%%%%%%%%%%%%%%%%%%%%%%%%%%%%%%55

Consider a filtration $F=(\sigma_1, \sigma_2, \dots, \sigma_m)$
 of  a simplicial complex $K$ obtained from a given filter function 
$f:K\to\mathbb{R}$.
If $\sigma_{i}$ completes a $p$-cycle ($p$ being the dimension of $\sigma_{i}$) when $\sigma_{i}$ is added to $F_{i-1}=(\sigma_1,\dots,\sigma_{i-1})$, then a $p$-homology class $\alpha$ {\it is born at time $f(\sigma_i)$}; otherwise, a $(p-1)$-homology class {\it dies at time $f(\sigma_i)$}.
 The difference between  the birth and death times of a homology class $\gamma$ is called its {\it persistence}, which quantifies the significance of a topological attribute. If $\alpha$ never
dies, we set its persistence to infinity. 

%%%%%%%%%%%%%%%%%%%%%%%%%%%%%

For a $p$-homology class that is born at time $f(\sigma_{i})$ and dies at time  $f(\sigma_{j})$, we draw a {\it bar} 
$[f(\sigma_{i}), f(\sigma_{j}))$
with endpoints  $f(\sigma_{i})$ and $f(\sigma_{j})$.
The set of bars $\{[f(\sigma_{i}), f(\sigma_{j}))\subset {\mathbb R}\}$
(resp.  points $\{(f(\sigma_{i}), f(\sigma_{j}))\in {\mathbb R}^2\}$) representing birth and death times of homology classes is called the {\it persistence barcode} $B(F)$ 
(resp. {\it persistence diagram} $dgm(F)$)
for the filtration $F$. 
Analogously,  fixed $i$, the set of bars (resp. points)
representing birth and death time of $i$-homology classes is called 
the {\it $i$-persistence barcode} 
(resp. the {\it $i$-persistence diagram}) 
for $F$.

\begin{figure}[t!]
\centering
\includegraphics[width=3 in]{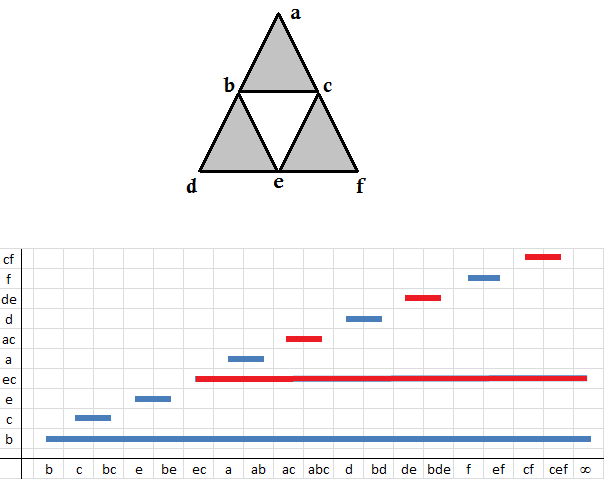} %12.5
\caption{An example of a persistence barcode obtained from a simplicial complex.}
 \label{persistent}
\end{figure}

For example, in Figure \ref{persistent}, the filtration 
$F=\{$ $b$, $c$ ,$bc$ ,$e$, $be$, $ec$, $a$, $ab$, $ac$, $abc$, $d$, $bd$, $de$, $bde$, $f$, $ef$, $cf$, $cef\}$ of the simplicial complex which consists of a composition of three triangles, can be read on the $x$-axis of the picture. 
Bars corresponding to the persistence of $0$-homology classes (i.e. the persistence of connected components) are colored in blue and bars corresponding to the persistence of $1$-homology classes (i.e., the persistence of tunnels)
 are colored in red.  Observe that only two bars survive until the end: 
 one corresponding to the connected component and  one corresponding to the tunnel of the simplicial complex.

%%%%%%%%%%%%%%%%%%%%%%%%%%%%%%%%%%%%%%%%%%%%%%%%%%

The bottleneck distance (see \cite[page 229]{compTopologyBook}) is classically used to compare two persistence diagrams $dgm(F)=\{a_1,\dots,a_k\}$ and $dgm(F')=\{a'_1,\dots,a'_{k'}\}$ for two  filtrations $F$ and $F'$ of, respectively, two finite simplicial complexes $K$ and $K'$. 
 The {\it bottleneck distance} between $dgm(F)$ and $dgm(F')$ is:
$$d_b(dgm(F),dgm(F'))=\min_{\gamma}\{\max_a\{||a-\gamma(a)||_{\infty}\}\}$$ 
 where $\gamma: dgm(F)\to dgm(F')$ is a bijection that can associate a point off the diagonal with another point on or off the diagonal, where {\it diagonal} is the set $D=\{(x,x)\}\subset {\mathbb R}^2$. For points $a=(x,y)\in F\cup D$ and $\gamma(a)=(x',y')\in F'\cup D$, the expression
$||a-\gamma(a)||_{\infty}$ means $\max\{ |x-x'|, |y-y'|\}$. 
Observe  that since  $K$ and $K'$ are finite then so are $dgm(F)$ and $dgm(F')$.
 Table \ref{botleneck} in page \pageref{botleneck} shows the bottleneck distance between the persistence diagrams pictured in Figure \ref{diagram}.

The following definitions are taken from \cite{chazal}.
Let $W$ and $W'$ be the   
vertex set of, respectively,  two simplicial complexes $K$ and $K'$. A {\it correspondence} 
$C:W \Rightarrow W'$ from $W$ to $W'$ is a subset of $W\times W'$ 
satisfying that  for any $v\in W$ there exists $v'\in W'$ such that $(v,v')\in C$ and, conversely,  for any $v'\in W'$ there exists $v\in W$ such that $(v,v')\in C$. 
Besides, for a subset  $\sigma$ of $W$, $C(\sigma)$ denotes the subset of $W'$ satisfying that a vertex $v'$ is in $C(\sigma)$ if and only if there exists a vertex $v\in \sigma$ such that $(v,v')\in C$.
Finally, given filter functions $f : K\to \mathbb{R}$
and $f': K'\to \mathbb{R}$, and the corresponding filtrations $F$ and $F'$, we say that
$C:W \Rightarrow W'$ is {\it $\epsilon$-simplicial  from $F$ to $F'$} if for any $t\in \mathbb{R}$ and  simplex $\sigma\in K$ such that $f(\sigma)\leq  t$, every simplex $\mu\in K'$ with vertices in $C(\sigma)$ satisfies that $f'(\mu)\leq t+\epsilon$. 
The transpose of $C$, denoted by $C^T$, is the image of $C$
through the symmetry map $(x,y)\to(y,x)$.

The following results will be used later in the paper.

%%%%%%%%%%%%%%%%%%%%%%%%%%%%%%%%%%%%%%%%%%%%%%%%%%%%%%%%%%%%%%%%

\begin{proposition}\label{prop:42}
{\cite[Proposition 4.2]{chazal}}
 Let $S$ and $T$ be filtered complexes with vertex sets $X$ and $Y$ respectively. If $C :
X \Rightarrow Y$ is a correspondence such that $C$ and $C^T$ are both $\epsilon$-simplicial, then together they
induce a canonical $\epsilon$-interleaving between $H(S)$ and $H(T)$, the interleaving homomorphisms
being $H(C)$ and $H(C^T )$.
\end{proposition}

The homology groups of each subcomplex in a filtered complex together with the  morphism between homology groups induced by the inclusion maps is an example of a $q$-tame module if all the subcomplexes are finite which is the case we deal with in this paper. 

\begin{theorem}\label{th:23}
{\cite[Theorem 2.3]{chazal}}
 If $U$ is a $q$-tame module then it has a well-defined persistence diagram
$dgm(U)$. If $U$, $V$ are $q$-tame persistence modules that are $\epsilon$-interleaved then there exists an
$\epsilon$-matching between the multisets $dgm(U)$ and $dgm(V)$. Thus, the bottleneck distance between
the diagrams satisfies the bound $d_b(dgm(U), dgm(V)) \leq\epsilon$.
\end{theorem}

\section{Topological signature for a periodic   motion}
\label{simpcompsec}

In this section,
we explain how to compute the topological signature for a  sequence of silhouettes. First, in Subsection \ref{simplicial}, a simplicial complex is built from the given sequence. Then, in Subsection \ref{filtration}, eight filtrations of the simplicial complex are computed in order to capture the movements that characterize the motion recorded in the sequence. In Subsection \ref{signat} we finally explain how to obtain the signature from the persistence barcodes for the given filtrations.
%As we said in the introduction, this signature has been used for human identification  \cite{lamar2012human}, gender classification \cite{leon2013gait},  carried object detection  \cite{ciarp2014}, monitoring human activities at distance  \cite{icpr2014}, 3D face expression recognition  \cite{yonghzhen} and  forehand and backhand strokes  performed by  a tennis player classification  \cite{MJ}. In \cite{icpr2016}, the signature is applied on the lower part of the body avoiding the variability in the upper body part. 

\subsection{From sequences of silhouettes  to simplicial complexes}\label{simplicial}

\begin{figure}[ht!]
\centering
\includegraphics[width=\textwidth]{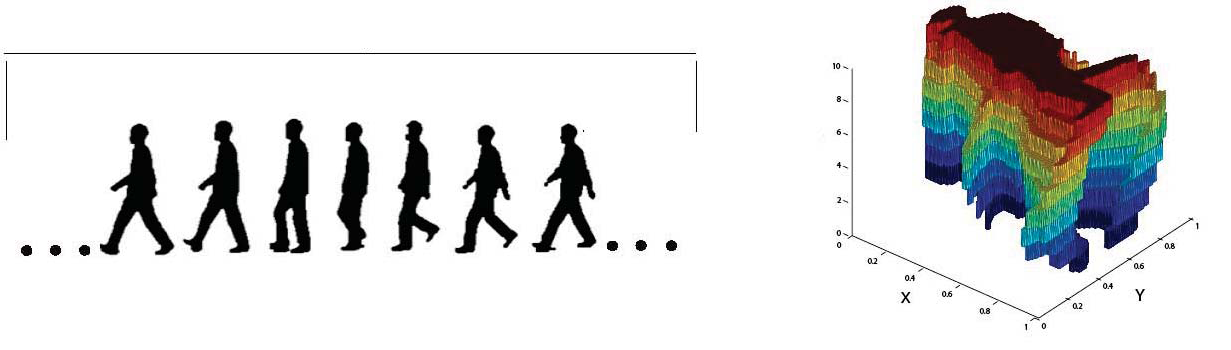} %
\caption{a) Left: sequence of gait silhouettes.
b) Right: associated simplicial complex $\partial K(I)$.}
\label{fig:dK}
\end{figure}

In this subsection we introduce the construction of the simplicial complex $\partial K(I)$ which models  the input  sequence, which is a sequence of silhouettes obtained from a periodic   motion sequence. 
See, for example, Figure \ref{fig:dK}.a.
In our previous papers \cite{lamar2012human,leon2013gait,ciarp2014,icpr2014,icpr2016}, we got the sequences from the background segmentation provided in CASIA-B dataset\footnote{http://www.cbsr.ia.ac.cn/GaitDatasetB-silh.zip} for gait silhouettes. 

\begin{figure}[ht!]
 \centering
 \includegraphics[width=2.5 in]{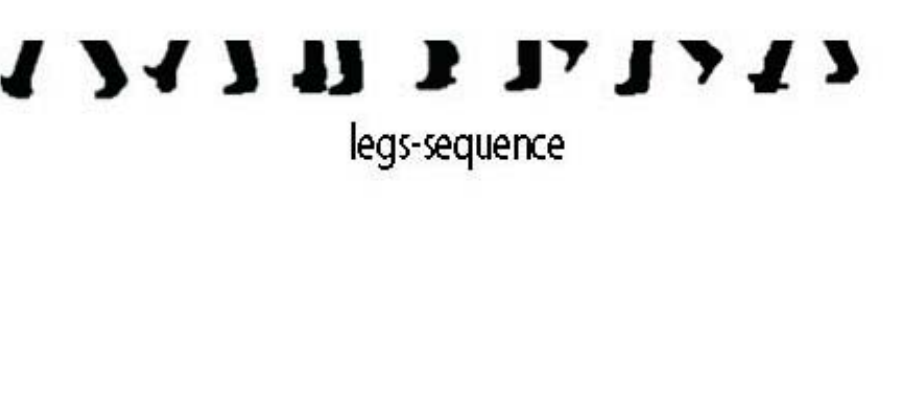} 
 \includegraphics[width=2 in]{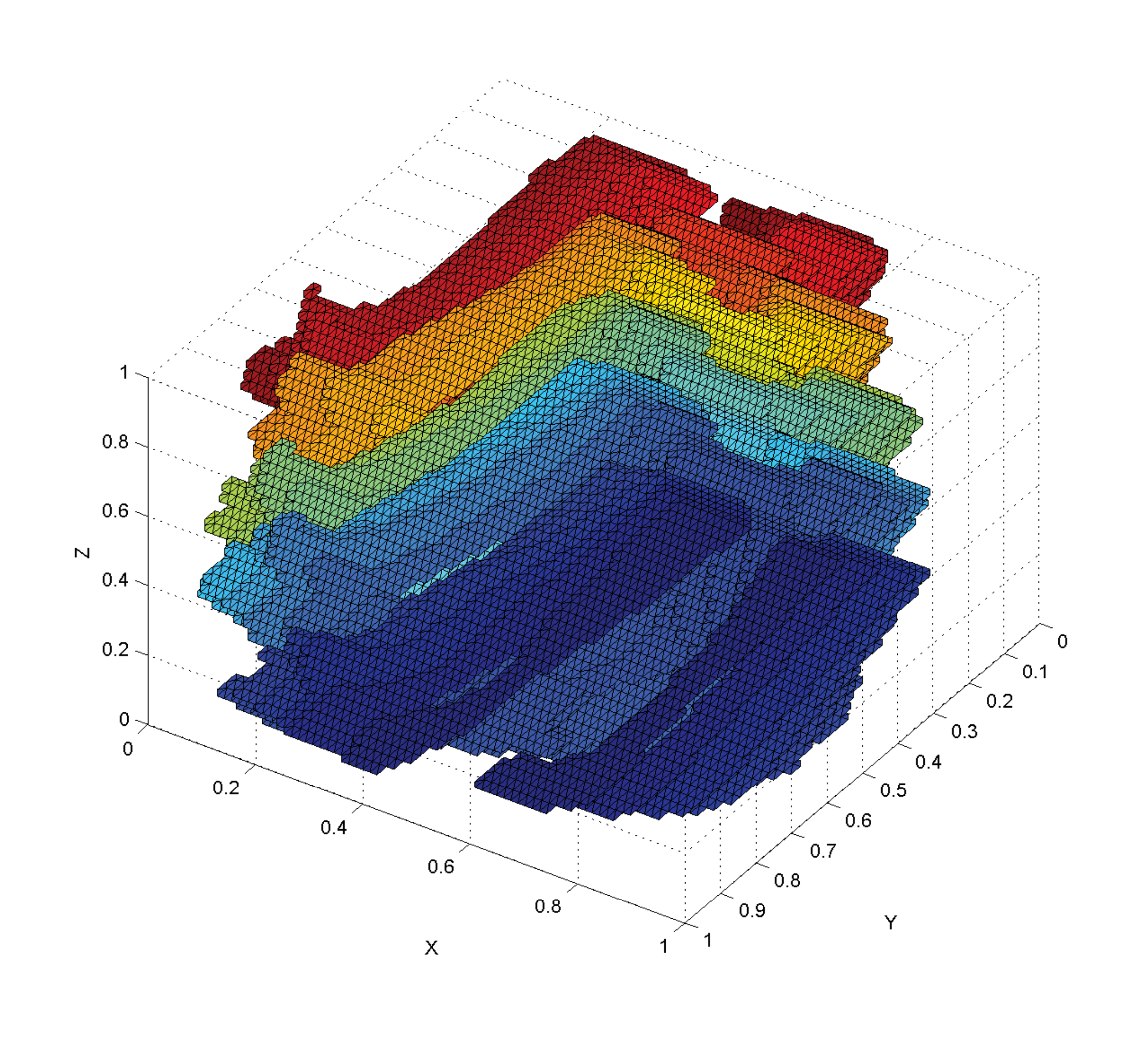} 
 \caption{a) Left: sequence of leg-silhouettes. b) Right: associated simplicial complex $\partial K(I)$.}
 \label{legs}
 \end{figure}

We  build a 3D binary  image $I = (\mathbb{Z}^3,B)$  by stacking $k$ consecutive silhouettes.  
The cubical complex $Q(I)$ associated to the 3D binary image $I$ is then computed.
The height of each silhouette is set to $1$ and the   width changes accordingly to preserve the original proportion between height and width. Besides,  
$z$-coordinates that represented the amount of silhouettes in the stack 
 is also set to $1$. 
Then, by construction, $x$-, $y$- and $z$-coordinates of the vertices in $Q(I)$ have their values in the interval $[0,1]$.  
 Finally, the squares that are faces of {\bf exactly one} cube in $Q(I)$ are divided into two triangles. These triangles together with their faces (vertices and edges) form the simplicial complex $\partial K(I)$.
 See Figure \ref{fig:dK}.b and Figure \ref{legs}.b.
 The different colors in the figures are used just to recognize each silhouette in the complex.

\subsection{Filtration of the simplicial complex $\partial K(I)$}\label{filtration}

The next step in our process is to compute filtrations of the previously computed simplicial complex $\partial K(I)$, in order to capture the movements recorded in the sequence.

\begin{figure}[ht!]
\centering
\includegraphics[width=\textwidth]{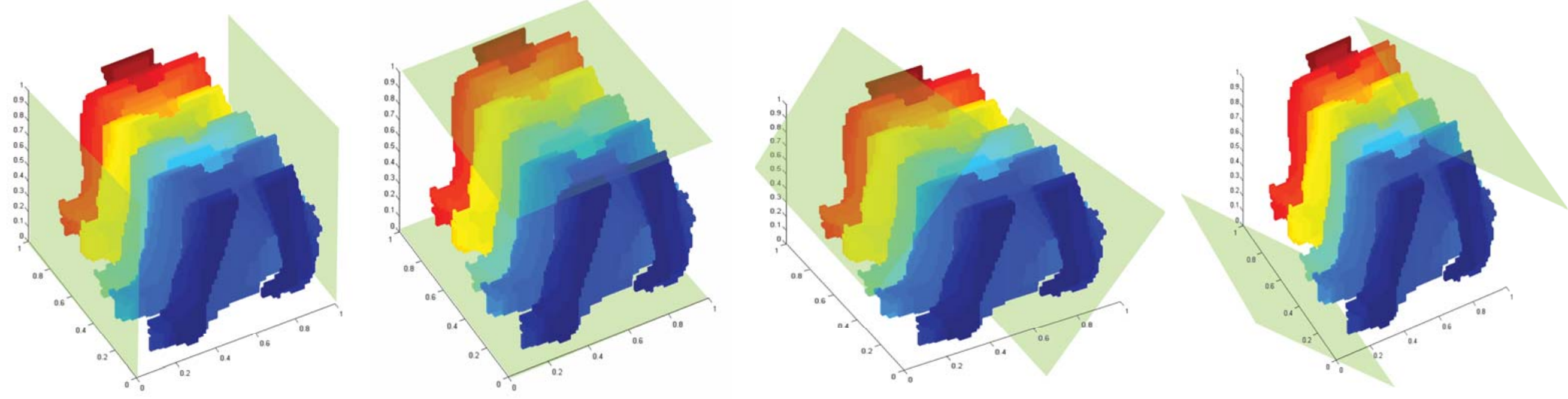} %
\caption{
From left to right: the eight planes
 used to compute the eight filtrations of $\partial K(I)$: two vertical, two horizontal  and four oblique planes. In this figure,  $\partial K(I)$ is the simplicial complex showed in Figure \ref{legs}.b.}
\label{planos2_img}
\end{figure}

Eight different filtrations of $\partial K(I)$ are computed
using, respectively, eight planes: two vertical planes ($x=0$ and $x=1$), two horizontal planes ($y=0$ and $y=1$) and four oblique planes ($x-y=1$, $y-x=1$, $x+y=0$ and $x+y=2$). See Figure \ref{planos2_img}.

More concretely, for each plane $\pi$, we define the filter
 function $f_{\pi}:\partial K(I)\rightarrow \mathbb{R}$ which assigns to each vertex of $\partial K(I)$ its distance to the plane $\pi$, and to any other simplex of $\partial K(I)$,  the greatest distance of its vertices to $\pi$. 
A filtration $\partial K_{\pi}$ 
of $\partial K(I)$ is then computed 
dictated by the filter function  $f_{\pi}$.

Observe that for a vertex $v\in \partial K(I)$, the value of 
$f_{\pi}(v)$ is:
(a)
less or equal than $1$ if $\pi$ is an horizontal o vertical plane; 
and  (b) less or equal than $\sqrt{2}$ if $\pi$ is an oblique plane.
Finally, observe that not any filtration involves $z$-coordinate. The reason for this is that our aim is to obtain a topological signature robust to the number of silhouettes (which was one of the weaknesses of previous approaches as mentioned in the introduction) used to compute the simplicial complex  $\partial K(I)$.

%%%%%%%%%%%%%%%%%%%%%%%%%%%%%%%%%%%%%

\subsection{Topological signature} \label{signat}

The final step in our process  is to compute the 
persistence barcode  for each filtration 
$\partial K_{\pi}$ 
associated to each plane $\pi$ (see  Figure \ref{planos2_img}).

%%%%%%%%%%%%%%%%%%%%%%%%%%%%%%%%%%%%%%%%

\begin{figure*}[t!]
\centering
\includegraphics[width=12cm]{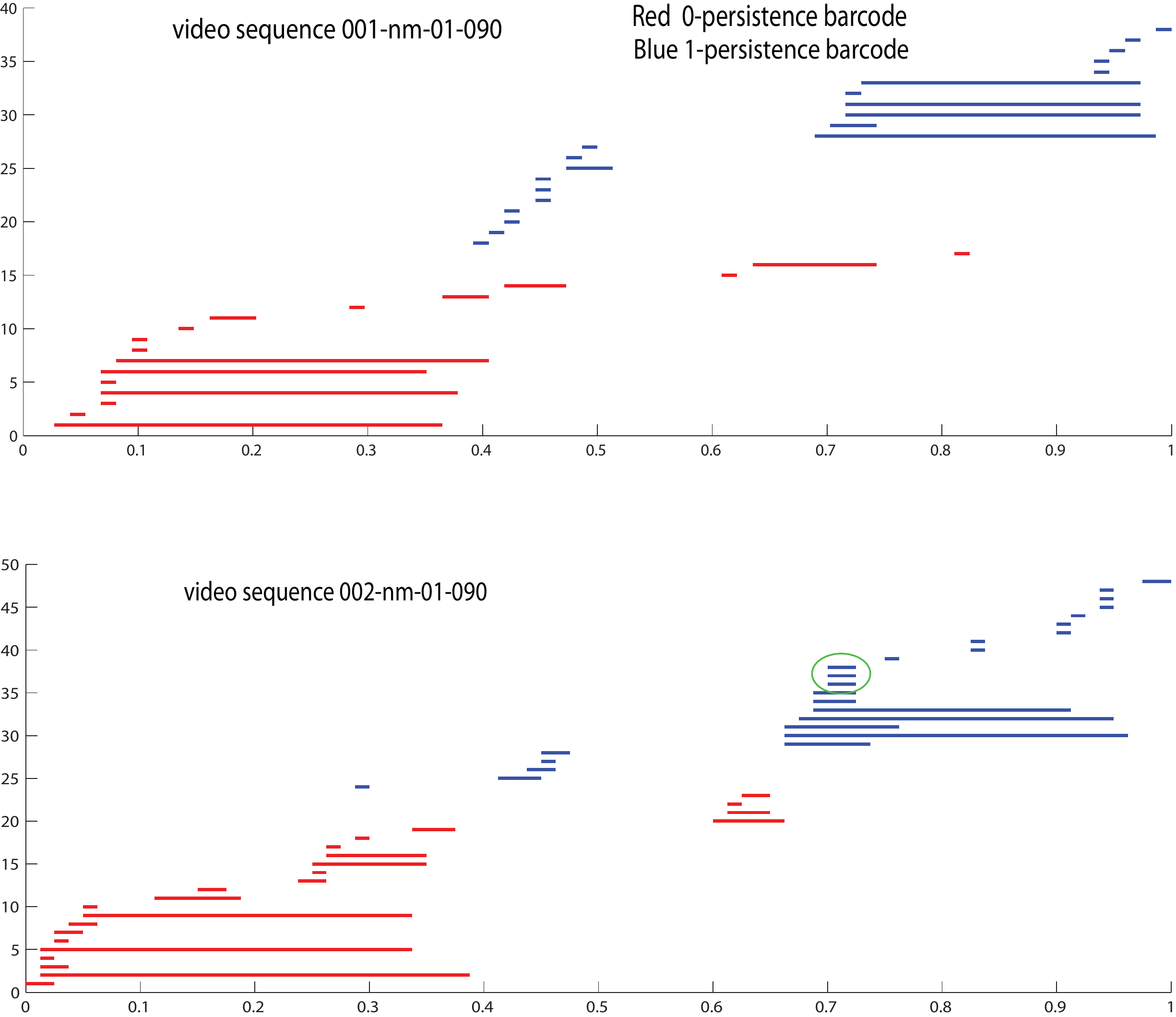} %12.5
\caption{Persistence barcodes for two filtrations obtained, respectively, from the sequences 001-nm-01-090 and 002-nm-01-090 of CAISA-B dataset. Horizontal axis represents the distance to the reference plane $x=0$.}
 \label{codebar}
\end{figure*}

%%%%%%%%%%%%%%%%%%%%%%%%%%%%%%%%%%

We only consider bars in the persistence barcode with length strictly greater than $0$. This way, we do not take into account any topological event $e$ that is born and dies at the same distance to the reference plane. This is not a problem in the sense that we could lose information, since that event $e$ will be captured using a different reference plane.

%%%%%%%%%%%%%%%%%%%%%%%%%%%%%%%%%%%%%%5

As an example, the persistence barcodes for two filtrations dictated by the distance to the plane $x=0$ and obtained, respectively, from  the video sequences 001-nm-01-090 and 002-nm-01-090 of the CAISA-B dataset,  
are shown in Figure \ref{codebar}. 
The set of red bars forms the $0$-persistence
barcode
and the set of blue bars forms the $1$-persistence
barcode. 
Notice the green circle showing topological features that are born and die at the same time.

%%%%%%%%%%%%%%%%%%%%%%%%%%%%%%%%%%%%%%%%%%%%%

Now,  for each plane $\pi$, the $0$- and $1$-persistence barcode    for the filtration $\partial K_{\pi}$ are explored according to a uniform sampling.
More concretely, given a positive integer $n$ (being $n=24$ in our experimental results, obtained by cross validation), we compute the value $h=\frac{k}{n}$, which represents the width of the ``window'' we use to analyze the persistence barcode, being $k$ the greatest distance of a vertex in $\partial K(I)$ to the plane $\pi$.
Since the distance to the plane $\pi$ has been normalized  then $k\leq \sqrt{2}$, so $h\leq \frac{\sqrt{2}}{24}$.

\begin{proc}\label{vector}
A vector  ${\cal V}_{\pi}^0$ (resp. ${\cal V}_{\pi}^1$)  of $2n$ entries is  constructed as follows.
For $s=0,\dots,n-1$: 
\begin{itemize}
\item[(a)]
entry $2s$ contains 
 the number of $0$- (resp. $1$-) homology classes that are born before $s\cdot h$ and persist or die after $s\cdot h$; 
 \item[(b)] 
 entry $2s+1$ contains 
 the number of $0$- (resp. $1$-) homology classes that are born in $s\cdot h$ or later and before $(s+1)\cdot h$.
 \end{itemize}
 \end{proc}

For example, suppose that a homology class is born and
dies within the interval $[s \cdot h,(s + 1)\cdot h)$. Then, in this case,
we add $1$ in entry $2s + 1$. On the other hand, suppose that a
homology class is born in the interval $[s \cdot· h,(s + 1)\cdot h)$ and
dies in $[t \cdot h,(t + 1) \cdot h)$ for  some $s,t$ such that
$ s < t \leq n$. Then, in this
case, we add $1$ in entry $2s+1$, and in entries $2j$ for $s < j \leq  t$.

Dividing the entries in two categories (a) and (b),  small details in the object are highlighted, which is crucial for distinguishing two different motions. For example,  let us suppose a scenario in which $m$ $0$-homology classes are born in   $[s\cdot h, (s+1)\cdot h)$ and persist or die at the end of $[(s+1)\cdot h, (s+2)\cdot h)$ and not any other $0$-homology class is born, persists or dies in these intervals. Then, we put $0$ in entries $2s$ and $2s+3$ of ${\cal V}_{\pi}^0$, and $m$ in entries $2s+1$ and  $2s+2$ of ${\cal V}_{\pi}^0$. On the other hand, let us suppose that $m$ $0$-homology classes are born and die in   $[s\cdot h, (s+1)\cdot h)$ and in $[(s+1)\cdot h, (s+2)\cdot h)$ and not any other $0$-homology class is born, persists or dies in these intervals. Then, we put $0$ in entries $2s$ and $2s+2$ of ${\cal V}_{\pi}^0$ and $m$ in entries $2s+1$ and $2s+3$ of ${\cal V}_{\pi}^0$. 
Therefore,  considering (a) and (b) separately, we can distinguish both scenarios.
 
Fixed a plane $\pi$, we then obtain two $2n$-dimensional vectors for $\partial K_{\pi}$, one for the $0$-persistence barcode and the other for the $1$-persistence barcode for the filtration  
$\partial K_{\pi}$.
Since we have eight planes, $\{\pi_1,\dots \pi_8\}$, and two vectors per plane,  $\{{\cal V}^0_{\pi_i}, {\cal V}^1_{\pi_i}\}:$ $i=1,\dots,8$, we have a total of sixteen $2n$-dimensional vectors which form the {\it topological signature for a periodic  motion sequence}.

Finally, to compare the topological signatures for two periodic motion sequences, we add up the angle between each pair of vectors in the  signatures. Since a signature consists of sixteen vectors, the best comparison for two sequences is obtained when the total sum is $0$ and the worst is $90^o\cdot16=1440^o$. 
Observe that in our previous papers \cite{lamar2012human,leon2013gait,ciarp2014,icpr2014}, we used the cosine distance\footnote{The cosine distance between two  vectors $v$ and $w$ is $\frac{v\cdot w}{||v||\cdot ||w||}$.}
to compare two given topological signatures. In that case, the best comparison for two sequences is obtained when the total sum is $16$ and the worst comparison when it is $0$. 

\begin{remark}\label{remark:angle}
We have  noticed that using the angle instead of the cosine to compare two topological signatures, the efficiency (accuracy) of gait recognition increases by $5\%$. This comparison is made in Table \ref{table1}.
The reason for this phenomena is that angle is more discriminative than cosine when the angle between vectors is close to zero\footnote{Recall that 
$\lim_{x\to 0}\frac{x}{1-\cos(x)}=\infty$.}. 
\end{remark}

  \section{Stability of the topological signature for a  periodic motion sequence}
\label{section:stability}

Once we have defined the topological signature for a periodic motion sequence, our aim  is to prove its stability under small perturbations on the  sequence (Theorem \ref{main}) and/or  under variations on the number of periods in the sequence (Theorem \ref{main1}).

The following technical statements will be used to prove Theorem \ref{main}.

\begin{proposition} \label{prop:previo2}
 Let 
 $F$ (resp. $F'$) be a filtration of a simplicial complex $K$ (resp. $K')$ dictated by a filter function $f:K\to \mathbb{R}$ (resp. $f':K'\to \mathbb{R}$).  
 Let $W$ (resp. $W'$) be the vertex set of $K$ (resp.  $K'$).
\\
If  $C:W\Rightarrow W'$ is a correspondence from $W$ to  $W'$  satisfying that 
$f'(v')\leq f(v)+\epsilon$ 
 %$d(v,v')\leq \epsilon$ 
 for every $(v,v')\in C$, 
then 
$$\mbox{$C:W\Rightarrow W'$ is   $\epsilon$-simplicial 
from $F$ to $F'$.
}$$
\end{proposition}

\noindent{\bf Proof.} 
Let $t\in \mathbb{R}$ and  $\sigma\in K$ such that $f(\sigma)\leq  t$. Then, every simplex $\mu\in K'$ with vertices in $C(\sigma)$ satisfies that $f'(\mu)\leq t+\epsilon$.
Then,   $C:W\Rightarrow W'$ is $\epsilon$-simplicial  from $F$ to $F'$ 
by definition.
\qed

\begin{proposition} \label{prop:previo}
Let 
 $F$ (resp. $F'$) be a filtration of a simplicial complex $K$ (resp. $K')$ dictated by a filter function $f:K\to \mathbb{R}$ (resp. $f':K'\to \mathbb{R}$).  
 Let $W$ (resp. $W'$) be the vertex set of $K$ (resp.  $K'$).
\\
If the correspondence  $C:W\Rightarrow 
W'$ is $\epsilon$-simplicial from $F$ to $F'$,  then 
$$d_b(dgm(F),dgm(F'))\leq \epsilon.$$
\end{proposition}

\noindent{\bf Proof.}
The statement is a direct consequence of  Proposition~\ref{prop:42} in page \pageref{prop:42} and Theorem~\ref{th:23}.
\qed

%%%%%%%%%%%%%%%%%%%%%%%%%%%%%%%%%%%%%%%%%%%%

In the following theorem, we prove, in terms of probabilities, that the topological signature is stable under small perturbations on the input data (i.e., the input sequence).

\begin{theorem} \label{main}
Let $I$ (resp. $I'$) be a 3D binary image. 
Let $W$ (resp. $W'$) be the vertex set of $\partial K(I)$ (resp.  $\partial K(I')$).
Let $\partial K_{\pi}$ (resp.  $\partial K'_{\pi}$) be the  filtration of
$\partial K(I)$ (resp.  $\partial K(I')$)
dictated by the distance function
$f_{\pi}:\partial K(I)\to \mathbb{R}$ 
(resp. $f'_{\pi}:\partial K(I')\to \mathbb{R}$)
to a given plane $\pi$.
Let ${\cal V}^j_{\pi}$ (resp.  
 ${\cal X}^j_{\pi}$), where $j=0,1$, be the 
two vectors obtained by applying Proc. \ref{vector} to the persistence barcode for $\partial K_{\pi}$ (resp.  $\partial K'_{\pi}$).
Let $m_i$ (resp. $m'_i$) be the  number of bars  in the $i$-persistence barcodes for the filtration  $\partial K_{\pi}$
(resp.  $\partial K'_{\pi}$).
\\
If  $C:W\Rightarrow W'$ is a correspondence from  $W$ to  $W'$ satisfying that 
$f'_{\pi}(v')\leq f_{\pi}(v)+\epsilon$ 
for every $(v,v')\in C$,
then 
$$\mbox{${\cal V}^i_{\pi} = {\cal X}^i_{\pi}$
with probability greater or equal than $\left(1-\frac{2(n-1)\epsilon}{k}\right)^{m_i+m'_i}$}$$
where:
\begin{itemize}
\item $k$ is the maximum  distance of a point in $\partial K_{\pi}$ to the plane $\pi$;
\item  $n$ is the number of subintervals (``windows") in which the interval $[0,k]$ is divided.
\end{itemize}
\end{theorem}

\noindent{\bf Proof.}
By Proposition \ref{prop:previo2},  we have that  $C:$ $W\Rightarrow W'$ is $\epsilon$-simplicial from $\partial K_{\pi}$ to  $\partial K'_{\pi}$.
By Proposition \ref{prop:previo}, we have that $$d_b(dgm(\partial K_{\pi}),dgm(\partial K'_{\pi}))\leq \epsilon.$$
Let $\gamma: dgm(\partial K_{\pi})\cup\{(x,x)\}\to dgm(\partial K'_{\pi})\cup\{(x',x')\}$ be the bijection such that 
$\max_a\{||a-\gamma(a)||_{\infty}\}=d_b(dgm(\partial K_{\pi}),dgm(\partial K'_{\pi}))$.
\\
Now, let  $a=(x,y)\in dgm(\partial K_{\pi})$ and $a'=(x',y')\in dgm(\partial K'_{\pi})$, being $x<y$ and $x'<y'$.
On the one hand, if $\gamma(a)=a'$, then $|x-x'|\leq \epsilon$ and $|y-y'|\leq \epsilon$.
On the other hand, if  $\gamma(a)$ lies in the diagonal,  then $|x-y|\leq \epsilon$. Similarly, if 
 $\gamma^{-1}(a')$ lies in the diagonal,  then $|x'-y'|\leq \epsilon$.
\\
Now, observe that  ${\cal V}^i_{\pi}$ can be different from ${\cal X}^i_{\pi}$ if there exists a point $(\alpha,\beta)$ in $dgm(\partial K_{\pi})$  or $dgm(\partial K'_{\pi})$ satisfying that $\alpha$ or $\beta$ belongs to $(sh-\epsilon,sh+\epsilon)$ for $h=\lfloor\frac{k}{n}\rfloor$ and  $s\in\{1,\dots,n-1\}$. 
Any of both situations ($\alpha$ or $\beta$ belongs to $(sh-\epsilon,sh+\epsilon)$) can occur with probability  $\frac{2(n-1)\epsilon}{k}$.
\\
Now, on the one hand,  suppose that  $\gamma(a)=a'$.  If
$x$ and $y$ do not belong to $(sh-\epsilon,sh+\epsilon)$
then $x'$ and $y'$ either.
On the other hand, suppose $\gamma(a)=(x',x')$ then $|x-y|\leq \epsilon$. If
$x\not\in(sh-\epsilon,sh+\epsilon)$ then $y$ either. Respectively, suppose  $\gamma^{-1}(a')=(x,x)$. If 
$x'\not\in(sh-\epsilon,sh+\epsilon)$ then $y'$ either.
\\
Therefore, 
${\cal V}^i_{\pi} = {\cal X}^i_{\pi}$
with probability greater or equal than $\left(1-\frac{2(n-1)\epsilon}{k}\right)^{m_i+m'_i}$.
 \qed

 As we will see next, the   result given in Theorem \ref{main} only makes sense when $\epsilon$ is ``small" enough.

 \begin{corollary}
 The probability of being
${\cal V}^i_{\pi} = {\cal X}^i_{\pi}$ tends to $1$ when $\epsilon$ tends to $0$.
Besides, such probability is non-negative when $\epsilon$ is  less or equal than $\frac{k}{2(n-1)}$ (recall that $h=\frac{k}{n}$ is the width of the``window'' we use to analyze the persistence barcode). 
 \end{corollary}
 
 \noindent{\bf Proof.}
 First, it is clear that if $\epsilon$ tends to $0$ then
 $\left(1-\frac{2(n-1)\epsilon}{k}\right)^{m_i+m'_i}$ tends to $0$.
 Second, $1-\frac{2(n-1)\epsilon}{k}\geq 0$ when  $\epsilon\leq \frac{k}{2(n-1)}$.
 \qed
 
 Take two 3D binary images $I$ and $I'$, 
 and a plane $\pi$ such that there exists a correspondence $C$ between the vertices of $\partial K(I)$ and $\partial K(I')$ satisfying that 
$f_{\pi}(v')\leq f_{\pi}(v)+\epsilon$ for any pair of  vertices $v\in\partial K(I)$ and $v'\in \partial K(I')$ matched by $C$.
Recall that, by construction, $f_{\pi}(v)$ and  $f'_{\pi}(v')$ are less or equal than $1$ (resp. $\sqrt{2}$) if $\pi$ is an horizontal or vertical plane
(resp. oblique plane).
Then $k\leq 1$ (resp. $k\leq \sqrt{2}$) if 
$\pi$ is an horizontal or vertical plane
(resp. oblique plane).
Now fix, for example,  $k=0.9$, $n=24$ and  $m_0=m'_0=20$.
Recall that  $n=24$ is the one used in our experimentation obtained by cross validation.
Half window size for this example is $0.9/(2\cdot 24)=0.01875$.
We have that
 ${\cal V}^0_{\pi}={\cal X}^0_{\pi}$ with probability greater or equal than $P$ for $P=0.814672814702977$ if  $\epsilon=0.0001$ and $P=0.979758005555892$ if  $\epsilon=0.00001$. 

Finally, the following result shows that the topological  signature does not depend on the number of periods in a given sequence.

\begin{theorem}\label{main1}
 The direction of the topological signature $\{{\cal V}^0_{\pi_i}, {\cal V}^1_{\pi_i}\}_{i=1,\dots,8}$ is independent on the number of periods the sequence contains.
\end{theorem}

\noindent{\bf Proof.}
Let $I_1$ be a stack of  consecutive silhouettes taken from a sequence of a periodic motion. Let $I_2$ be obtained by 
stacking twice the silhouettes of $I_1$. That is, for any point $v=(x,y,z)$ in $I_1$ there exist exactly two points $v=(x,y,z)$ and $v'=(x,y,z+k)$ in 
$I_2$ for a fixed $k$.
Let $\partial K(I_1)$  (resp. $\partial K(I_2)$ ) be the simplicial complex obtained from  $I_1$ (resp. $I_2)$).
Let $\pi$ be  the referred plane,  as one of the eight planes considered in this paper, used to compute the respective  filtrations
$\partial K^1_{\pi}$ and $\partial K^2_{\pi}$.
Since all the planes considered in this paper are perpendicular to plane $z=0$, then if the distance of  a vertex $v$ in $\partial K(I_1)$ to the plane $\pi$ is $d$, then the distance  
of vertices $v$ and $v'$ in $\partial K(I_2)$ to the plane $\pi$ is also $d$. Therefore, for any bar in the persistence barcode associated to the filtration $\partial K^1_{\pi}$, there exist exactly two bars in  the persistence barcode associated to the filtration $\partial K^2_{\pi}$ having the same birth and death times. Consequently, the vector corresponding to the topological signature for $I_2$ is exactly twice the vector corresponding to the topological signature for $I_1$.
\qed

\begin{figure*}[h!]
\centering
\includegraphics[width=\textwidth]{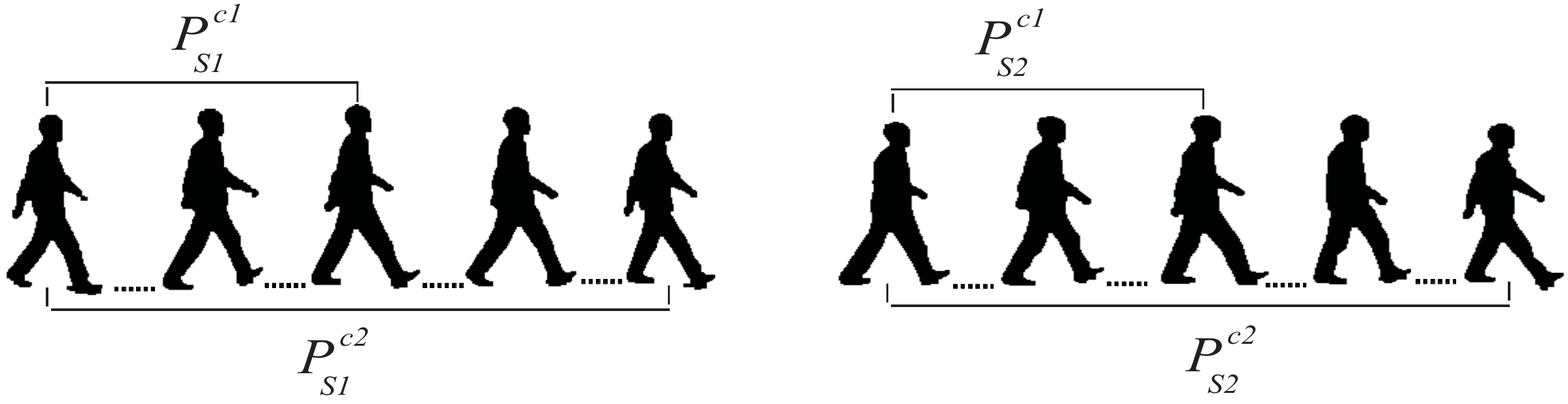} 
\caption{The silhouettes   extracted from two  gait sequences $S_1$ and $S_2$  of the same person.}
 \label{squence}
\end{figure*}

 \begin{figure*}[h!]
\centering
\includegraphics[width=\textwidth]{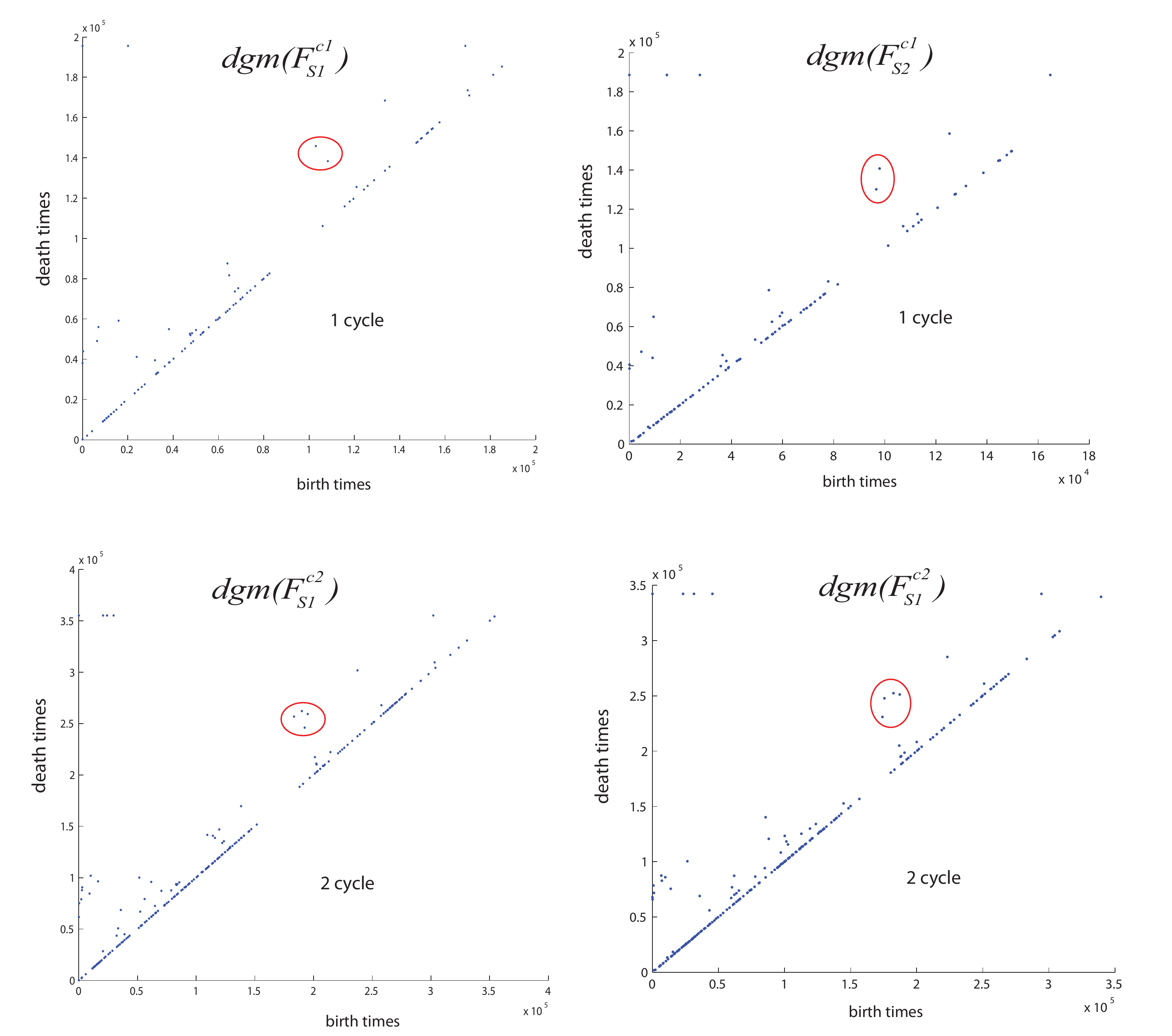} %12.5
\caption{Top: the $0$-persistence diagrams $dgm(F_{S1}^{c1})$ and $dgm(F_{S2}^{c1})$ obtained from sequences $S1$ and $S2$ 
 containing one gait cycle. Bottom: the $0$-persistence diagrams $dgm(F_{S1}^{c2})$ and $dgm(F_{S2}^{c2})$ obtained from   sequences $S1$ and $S2$
 containing   two gait cycles. The two sequences correspond to the same person.}
 \label{diagram}
\end{figure*}
 
For instance, let $S1$ and $S2$ be the silhouettes extracted from  two gait sequences of the same person from CASIA-B dataset, both $S1$ and $S2$ having  two gait cycles. 

Let $P_{S1}^{c1}$ and $P_{S2}^{c1}$ 
(resp.  $P_{S1}^{c2}$ and $P_{S2}^{c2}$) 
be the silhouette sequences of exactly one
gait cycle 
(resp. two gait cycles)
on $S1$ and $S2$.
See Figure \ref{squence}. 
Then, use the same reference plane $\pi$ to obtain the four filtrations 
$F_{S1}^{c2}$, $F_{S2}^{c2}$, $F_{S1}^{c1}$ and $F_{S2}^{c1}$ for the simplicial complexes associated to the 3D binary images obtained from the   sequences  $P_{S1}^{c2}$, $P_{S2}^{c2}$, $P_{S1}^{c1}$ and $P_{S1}^{c1}$, respectively.
The $0$-persistence diagrams  $dgm(F_{S1}^{c2})$, $dgm(F_{S2}^{c2})$, $dgm(F_{S1}^{c1})$ and $dgm(F_{S2}^{c1})$ are  showed in Figure \ref{diagram}. We can observe that the   diagrams 
 $dgm(F_{S1}^{c2})$ and $dgm(F_{S2}^{c2})$ have  approximately the double of persistent points than the   diagrams 
 $dgm(F_{S1}^{c1})$ and $dgm(F_{S2}^{c1})$ (look at  the area inside of the red circles  in Figure \ref{diagram}). Since the topological  signature is computed using ``windows'' in the persistence barcode (or equivalent, in the persistence diagram), then the modules of the topological signature obtained from the persistence diagrams  $dgm(F_{S1}^{c2})$ and $dgm(F_{S2}^{c2})$ is approximately the double of the modules of the topological  signature obtained from the persistence diagrams  $dgm(F_{S1}^{c1})$ and $dgm(F_{S2}^{c1})$. Nevertheless, the direction remains approximately the same. Observe that we do not have exactness since we do not stack the same silhouettes twice, but we take one or two cycles from a sequence, so small errors and variations may appear.

 \begin{table*}[ht!]
\centering
\caption{Cosine distance between the topological signatures obtained from the  persistence diagrams showed in Figure \ref{diagram}.}
\begin{tabular}{c|c c c }
\hline
 & $dgm(F_{S2}^{c1})$ & $dgm(F_{S2}^{c2})$ \\
\hline
% & &\\
 $dgm(F_{S1}^{c1})$ & 0.985 &  0.987 \\
 % & &\\
 $dgm(F_{S1}^{c2})$  & 0.981 &  0.990 \\
\hline
\end{tabular}
\label{coseno}
\end{table*}

 \begin{table*}[ht!]
\centering
\caption{Bottleneck distance between persistence diagrams showed in Figure \ref{diagram}.}
\begin{tabular}{c| c c c }
\hline
 & $dgm(F_{S2}^{c1})$ & $dgm(F_{S2}^{c2})$ \\
\hline
% & &\\
 $dgm(F_{S1}^{c1})$ & 855727 &  1319872 \\
%  & &\\
 $dgm(F_{S1}^{c2})$  & 2273559 &  5446584 \\
\hline
\end{tabular}
\label{botleneck}
\end{table*}

In Table  \ref{coseno} we show the results of the comparison between the topological signatures obtained from the persistence diagrams 
$dgm(F_{S1}^{c2})$,
 $dgm(F_{S2}^{c2})$, $dgm(F_{S1}^{c1})$ and
 $dgm(F_{S2}^{c1})$
using the  cosine distance. Observe that, in all the cases, the cosine distance is almost $1$ (i.e., all the vectors have almost the same direction), which makes sense since all the gait sequences correspond to the same person and the corresponding filtrations are computed using the same reference plane.
Finally, in Table \ref{botleneck} we show that 
 if we consider the classical bottleneck distance to compare the different persistence diagrams, we obtain different results depending on the number of gait cycles we consider to compute the topological signature. Therefore, the comparison using bottleneck distance does not provide useful information in this case.

\begin{figure*}[h!]
\centering
\includegraphics[width=\textwidth]{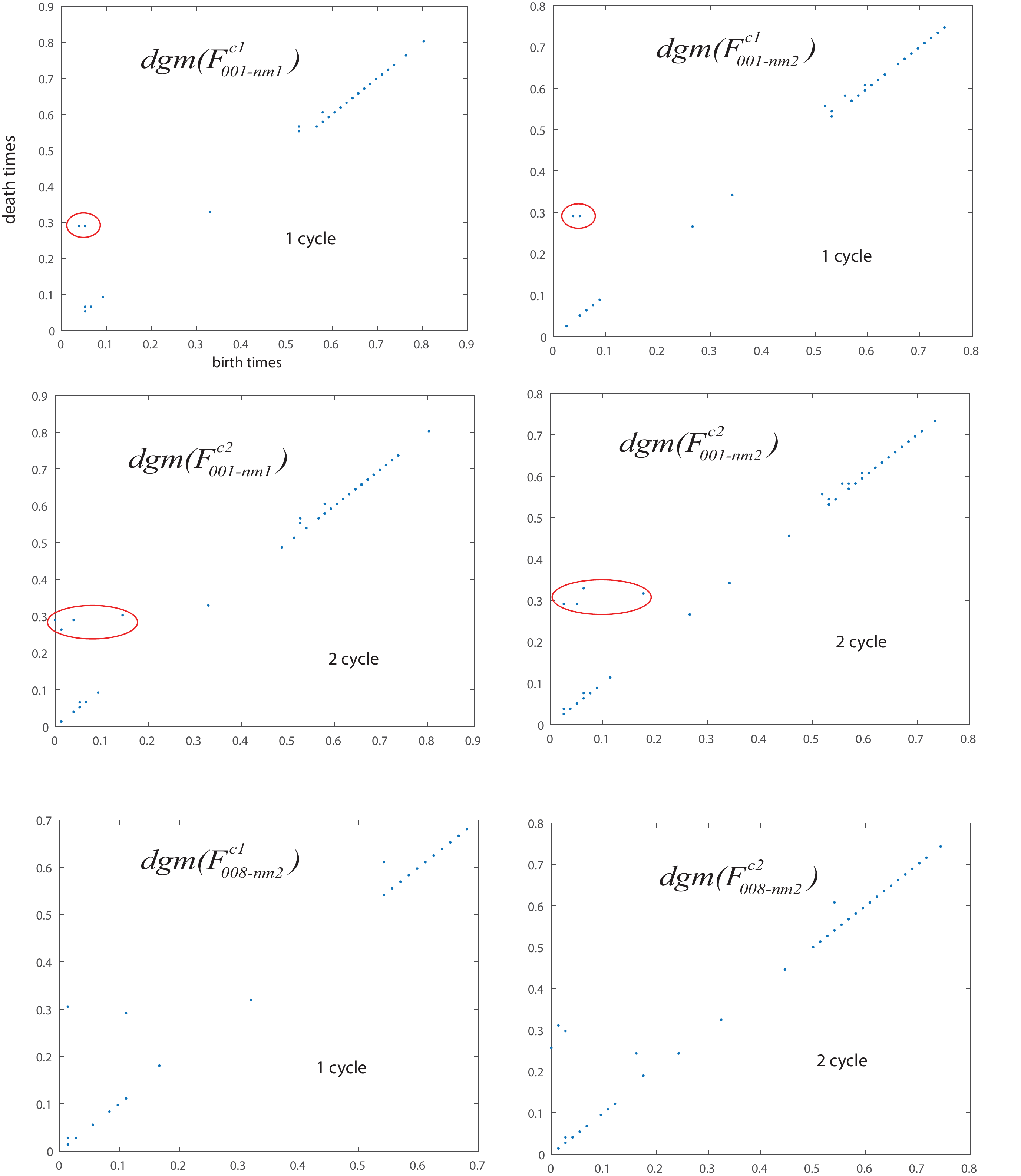} %12.5
\caption{Top: the two $0$-persistence diagrams 
 $dgm(F_{001-nm1}^{c1})$, $dgm(F_{008-nm2}^{c1})$. 
 Middle: 
 the two $0$-persistence diagrams 
 $dgm(F_{001-nm1}^{c2})$, $dgm(F_{008-nm2}^{c2})$.
 Bottom: 
  the two $0$-persistence diagrams 
 $dgm(F_{008-nm1}^{c1})$, $dgm(F_{008-nm2}^{c2})$.}
 \label{diagram1}
\end{figure*}

We now repeat the experiment for gait sequences obtained from two different persons and  considering only the lowest fourth part of the body silhouettes as in \cite{icpr2016}. 
Let 001-nm1 and 001-nm2 be two gait sequences of person 001 and let 008-nm2 be a gait sequence of person 008 taken from CASIA-B dataset. 
Let $\partial K(I_{001-nm1}^{ci})$,
 $\partial K(I_{001-nm2}^{ci})$ and $\partial K(I_{008-nm2}^{ci})$
 be the simplicial complexes associated to the 
 silhouettes sequences of exactly $i$ gait cycles on 001-nm1, 
001-nm2 and 008-nm2, for $i=1,2$. 
 
A fixed reference plane $\pi$  is used to obtain the $0$-persistence diagrams $dgm($ $F_{001-nm1}^{ci})$, $dgm(F_{001-nm2}^{ci})$ and $dgm($ $F_{008-nm2}^{ci})$
of the simplicial complex associated to 
$\partial K(I_{001-nm1}^{ci})$,
 $\partial K(I_{001-nm2}^{ci})$ and $\partial K(I_{008-nm2}^{ci})$,
 respectively. Observe in 
 Figure \ref{diagram1} that the persistence diagrams obtained from two gait cycles have twice as many points as the persistence diagrams obtained from one gait cycle. This can be noticed in the area inside of the red circles  in Figure \ref{diagram1}.
 
 \begin{table*}[ht!]
\centering
\caption{Bottleneck distance between persistence diagrams according to Figure \ref{diagram1}.}
$$\begin{array}{c| c c }
 &\, dgm(F_{001-nm2}^{c1}) \,&\, dgm(F_{001-nm2}^{c2}) \\
\hline
 dgm(F_{001-nm1}^{c1}) \,& 0.013 &  0.120 \\
 dgm(F_{001-nm1}^{c2})  \,& 0.125 &  0.040 \\
\hline
\end{array}$$
$$\begin{array}{c| c c }
 &\, dgm(F_{008-nm2}^{c1}) \,&\, dgm(F_{008-nm2}^{c2})\\
\hline
 dgm(F_{001-nm1}^{c1}) \, & 0.06 & 0.128\\
 dgm(F_{001-nm1}^{c2})  \, & 0.125 & 0.059\\
\hline
\end{array}$$
\label{botleneck1}
\end{table*}
 
\begin{table*}[ht!]
\centering
\caption{Cosine distance between the persistence diagrams showed in Figure \ref{diagram1}.}
$$\begin{array}{c| c c }
 &\, dgm(F_{001-nm2}^{c1}) \,&\, dgm(F_{001-nm2}^{c2}) \\
\hline
dgm(F_{001-nm1}^{c1}) \,& 0.944 & 0.931 \\
 dgm(F_{001-nm1}^{c2})  \,& 0.929 &  0.953 \\
\hline
\end{array}$$
$$\begin{array}{c| c c }
&\, dgm(F_{008-nm2}^{c1}) \,&\, dgm(F_{008-nm2}^{c2})\\
\hline
dgm(F_{001-nm1}^{c1}) \,& 0.739 & 0.886\\
 dgm(F_{001-nm1}^{c2})  \, & 0.832 & 0.905\\
\hline
\end{array}$$
\label{cosine1}
\end{table*}

In Table \ref{botleneck1} (resp. Table \ref{cosine1}),  we show the results for the comparison between the  persistence diagrams pictured in Figure \ref{diagram1}, using bottleneck distance (resp. cosine distance). 
The results show that the value of the bottleneck distance increases with the number of gait cycles, regardless of whether  we compare sequences of the same person. However, the values of the cosine distance between the topological signature for sequences of the same person are similar regardless of whether we consider different number of cycles in the sequence. 
%Furthermore, observe that the cosine distance between the topological signature of two sequences, of the same person, with two cycles,  is closer to one than if we only consider one cycle.

%%%%%%%%%%%%%%%%%%%%%%%%%%%%%%%%%%%%%%%%%%%%%%%%%%%%%%%%%%%%%%%%%%%%%%%%%%%%%%%%%%%%%%%%5

\section{Experimental Results}\label{exper}

In this section we show several experiments to support our approach. First,  results when applying our method on the CASIA-B dataset\footnote{http://www.cbsr.ia.ac.cn/GaitDatasetB-silh.zip} for gait recognition are given. Second, we compare different periodic motions such as jump, jack movement, run, skip or walk applying our approach to the
dataset of motion actions provided by L. Gorelick et al\footnote{http://www.wisdom.weizmann.ac.il/~vision/SpaceTimeActions.html}. Finally,  our method is evaluated on the OU-SIRT-B dataset\footnote{http://www.am.sanken.osaka-u.ac.jp/BiometricDB/GaitTM.html} for gait recognition when people use different  clothing types.
In all the experiments, the dataset is divided into two sets: the training set and the test set.
The topological signature for each person is obtained as  the average of the topological signatures computed from each of the samples of such person in the training dataset.

The CASIA-B dataset contains samples for each of the 11 different angles from which each of a total of 124 people is recorded. For each angle, there are 6 samples per person walking under natural conditions i.e.,  without carrying a bag or wearing a coat (CASIA-Bnm), 2 samples per person  carrying some sort of bag (CASIA-Bbg) and 2 samples per person wearing a coat (CASIA-Bcl). 
The CASIA-B dataset provides, for each sample, its background segmentation (called sequence). Summing up,  there are 10 sequences (6 from CASIA-Bnm, 2 from CASIA-Bbg and 2 from CASIA-Bcl) per person-angle.
We carried out two different experiments using the CASIA-B dataset.

\begin{table*}[ht!]
\centering
\caption{Accuracy (in $\%$) using a training dataset consisting of samples under natural 
conditions (i.e., without carrying a bag or wearing a coat).}
\scalebox{1}{
\begin{tabular}{ccccc}
\hline
Methods & CASIA-Bbg & CASIA-Bcl & CASIA-Bnm & Average
\\
\hline
Tieniu.T \cite{bb65346}  & 52.0 & 32.73 & 97.6 & 60.8\\
Khalid.B \cite{bashir2010gait} & 78.3 & 44.0 & 100 & 74.1\\
Singh.S \cite{singh2009biometric}& 74.58 & 77.04 & 93.44 & 81.7\\
Imad.R et al. \cite{rida2016gait}& 81.70 & 68.80 & 93.60 & 81.40  \\
Lishani et al. \cite{lishani2014haralick}& 76.90 & 83.30 & 88.70 & 83.00 \\
\textbf{Our Method }      & &  &  & \\
using cosine       & \textbf{80.5} & \textbf{81.7} & \textbf{92.4} & \textbf{84.9}\\
using angle       & \textbf{84.2} & \textbf{87.6} & \textbf{94.1} & \textbf{88.6}\\
\hline
\end{tabular}
\label{table1}}
\end{table*}

In the first experiment, from the total of 10 sequences per person-angle, we  used four sequences (taken from CASIA-Bnm dataset) to train, and the rest to test. Our results for lateral view (90 degrees) are shown in Table \ref{table1}, where we took the cross validation average ($\binom {6} {4}$  = 15 combinations) of accuracy at rank 1 from the candidate list. The experiment was carried out using only the legs of the body silhouette 
as in \cite{icpr2016}. As we have previously said in Remark~\ref{remark:angle}, we obtain better results when using the angle between vectors, instead of the cosine distance, to compare topological signatures.

\begin{figure}[h!]
\centering
\includegraphics[width=3.5 in]{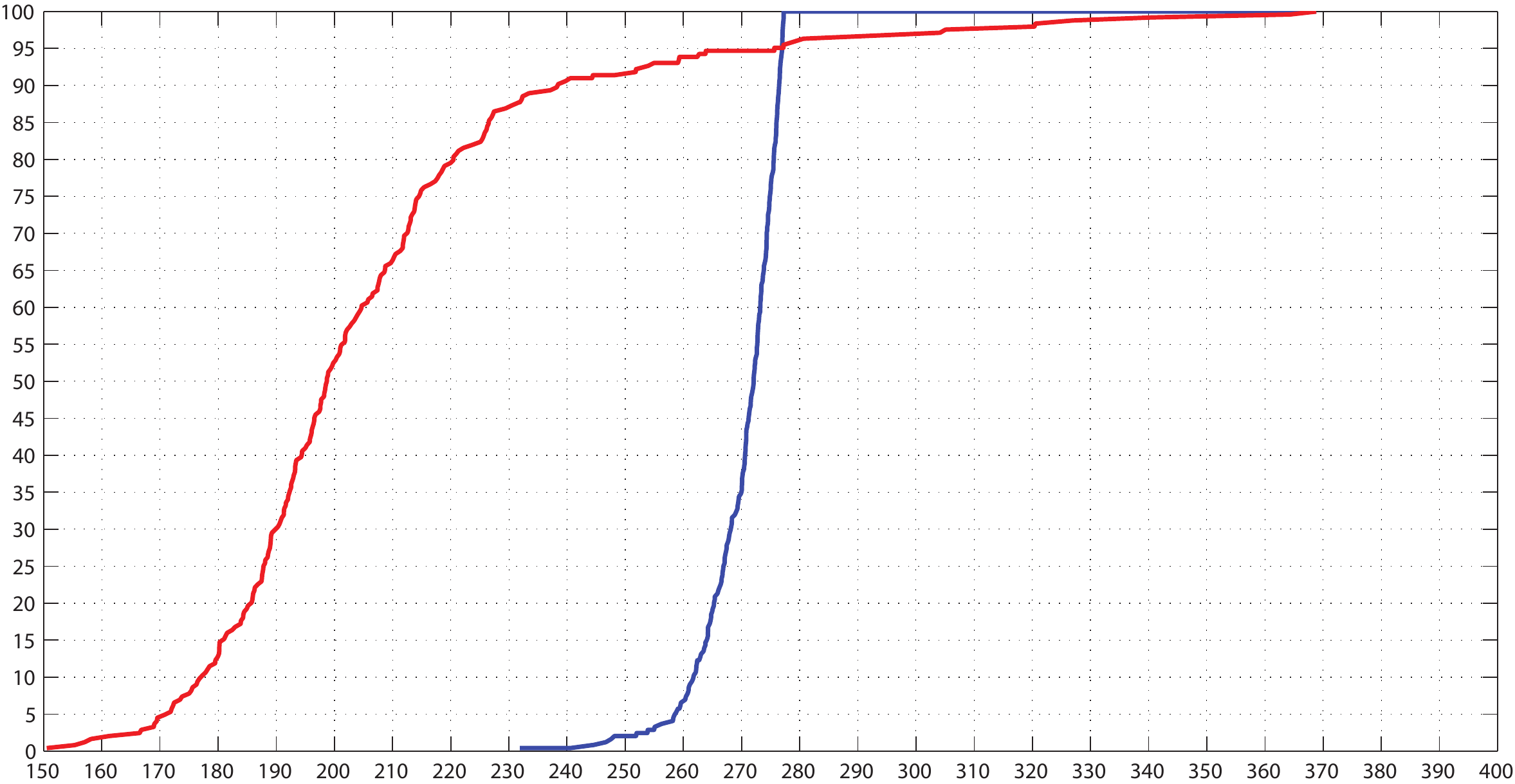} %
\caption{
Each point of the red curve (resp. blue curve) represents the percentage  of  values  of the TP set 
(resp. TN set)
lower than a threshold. For example, $92.6\%$ of the values of the TP set are  smaller than $253.8$, since the point $(253.8,92.6)$ belongs to the  red curve.
}
\label{unnamed}
\end{figure}

In the second experiment, we followed the protocol used in \cite[Section 5.3]{bashir2010gait}. This way, we considered a mixture of natural, carrying-bag and wearing-coat sequences, since it models a more realistic situation where persons do not collaborate while the samples are being taken. Specifically 
for lateral view (90 degrees),
from the ten sequences per person, 
six sequences were used  to train (four without carrying a bag or wearing a coat, one carrying a bag  and one wearing a coat) and  the rest was used for the test.  
Using this training dataset we generated $123$ topological signatures, one for each person in the dataset. 
 We must point out that person labeled as $005$ in CAISA-B was removed from the experiment due to extremely poor quality.
We used the remaining sequence of each person carrying a bag and the one wearing a coat for testing. This gave us $246$ sequences for testing: $123$ persons times $2$ sequences per person.
Now: 
\begin{itemize}
\item[(1)] For each person, we compared its previously computed topological signature with the two topological signatures obtained from each of the two test sequences. This way we obtain a set, called True Positive (TP) that contains $246$ comparison values.
\item[(2)]  For each person, we compared its previously computed topological signature
with the  two signatures obtained from each of the two test sequences of a different person.
This way we obtain a set,
 called True Negative (TN) that contains $123\cdot (2\cdot 122)=30012$ values.
\end{itemize}
We  restrict the TN sets to the $246$ smallest values, in order to balance the  TP and TN sets. 

Observe Figure \ref{unnamed} in which 
the $y$-axis  
represents percentages 
and the $x$-axis represents thresholds.
Each point of the red curve (resp. blue curve) shows the percentage  of  values of the TP set 
(resp. TN set)
lower than a threshold. 
Table \ref{table2} shows the  accuracy of the results using the angle between vectors  to compare two topological signatures.
In Table \ref{table1}  and Table \ref{table2} we can observe that  the best result of our method is obtained  for the set of natural sequences (CASIA-Bnm) and the worst  for the set of persons carrying bags.  This is due to that bags can affect on the lowest fourth part of the body silhouette (see  Figure \ref{afect_bag}). Moreover, the weight of the bag can change the dynamic of the gait. Nevertheless, accuracy results in all cases are greater that $80\%$. 
Besides, 
the features obtained from the lowest fourth part of the body silhouette gave an accuracy for the sequences of persons walking under natural conditions of $94.1\%$, which only decreases $3.9\%$ with respect to our previous paper \cite{lamar2012human} in which we used the whole body silhouette ($98.0\%$) and only considered persons walking under natural condition both for training and testing. This confirms that the highest information in the gait is in the motion of the legs, which supports the results given in \cite{bashir2010gait}.
As we can see in Table \ref{table1} and Table \ref{table2},  our method outperforms previous methods for gait recognition with or without carrying a bag or wearing a coat. 
Besides,  as we can see in Table \ref{table2}, the algorithm explained in \cite{bashir2010gait} decreases considerably the accuracy obtained by training mixing   the  natural, carrying-bag and wearing-coat sequences. On the contrary, our algorithm improves the accuracy for the whole test set  outperforming in more that 35\% the results given in  \cite{bashir2010gait}.
Comparing Table \ref{table1} and  Table \ref{table2}, we can as well arrive to the conclusion that training with more heterogeneous data gives to our method a more powerful representation for the classification step.

\begin{table*}[ht!]
\centering
\caption{Accuracy (in $\%$) using a training dataset consisting of samples under
different conditions (natural-walking, carrying-bag and wearing-coat).}
\scalebox{1}{
\begin{tabular}{c c c c c}
\hline
Methods & CASIA-Bbg & CASIA-Bcl & CASIA-Bnm & Average\\
 %&  & &   \\
\hline
Khalid.B \cite{bashir2010gait} & 55.6 & 34.7 & 69.1 & 53.1\\
\textbf{Our Method}  & \textbf{92.3} & \textbf{94.3} & \textbf{94.7} &  \textbf{93.8}\\
\hline
\end{tabular}
\label{table2}}
\end{table*}

Another experiment
was carried out using the  motion action dataset provided by L. Gorelick et al\footnote{http://www.wisdom.weizmann.ac.il/~vision/SpaceTimeActions.html}.
The aim of this experiment is to see how discriminative is our topological signature between different periodic motions.

\begin{figure}[h!]
\centering
\includegraphics[width=3.5 in]{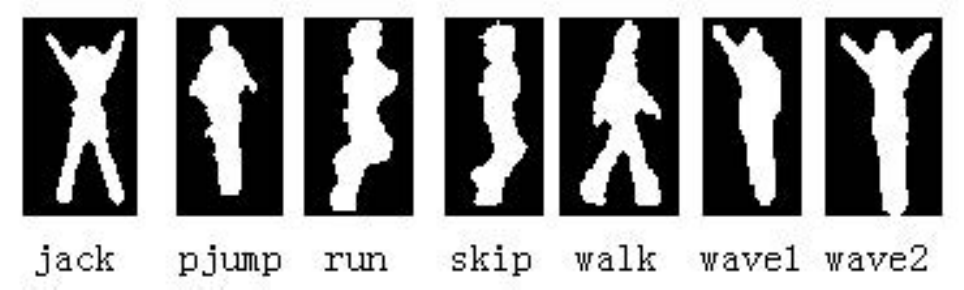} %
\caption{Silhouette samples taken from the dataset provided by L. Gorelick et al.}
\label{actions}
\end{figure}

This dataset has 7 actions, 8 samples by action, taken from 9 different persons (56  sequences in total). Figure \ref{actions} shows silhouette samples of these actions. To know the performance of our topological signature, we made several experiments with different number of samples in the training and test dataset. Concretely, let $a-b$ denote the experiment which consists of
$a$ samples per person  to train and $b$ to test. This way, we made $7$ different experiments: (A) $1-7$, (B) $2-6$, (C) $3-5$, (D) $4-4$, (E) $5-3$, (F) $6-2$, and (G) $7-1$. For instance, for A) $1-7$, we have 7 samples to train,  1 per action, and  49 samples to test,  7 per action.

\begin{table*}[ht!]
\centering
\caption{Total Accuracy (in \%) according to the dataset provided by L. Gorelick et al. for each of the  7 experiments considered.}
\scalebox{1}{
\begin{tabular}{c| c c c c c c c c}
\hline
& A & B& C& D& E& F & G \\ 
&1-7 & 2-6& 3-5& 4-4& 5-3& 6-2 & 7-1 \\
\hline
Rank 1& 74.6 & 82.1 & 85.7 & 86.8 & 88.9 & 90.1 & 90.1  \\ 
Rank 2& 89.3 & 96.1 & 97.1 & 97.8 & 98.5 & 98.1 & 98.7 \\ 
\end{tabular}
\label{all_accucary}}
\end{table*}

\begin{figure}[h!]
\centering
\includegraphics[width=1.5 in]{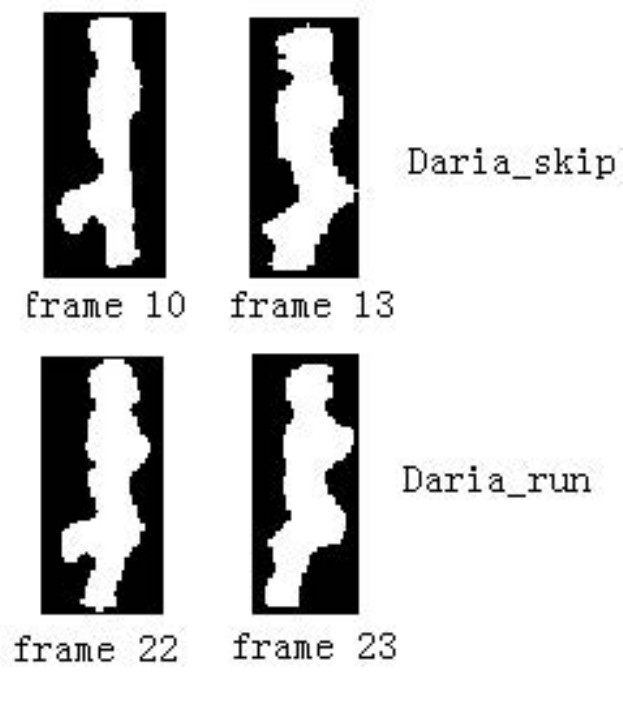} %
\caption{Similitude between skip and run action poses. Top: Daria-skip silhouettes. Bottom: Daria-run silhouettes.}
\label{confusion_skip_run}
\end{figure}

The samples to train were taken randomly in 100 iterations in order to compute the average accuracy. Table \ref{all_accucary} shows  the accuracy reached among all actions.  Besides, Table \ref{accucary_by_actions} shows the independent accuracy reached by each action. Table \ref{confusion_matrix-1} and 
Table  \ref{confusion_matrix-2}
show the confusion matrix for each experiment. We can see the confusion between  run and skip, which present similarities in their dynamics (as we can see in Figure \ref{confusion_skip_run}, there exists similitude in run and skip posses).

\begin{table*}[ht!]
\centering
\caption{Accuracy (in \%) reached by  action, according to the dataset provided by L. Gorelick et al. for each of the  7 experiments considered.}
\scalebox{1}{
\begin{tabular}{c| c c c c c c c c}
\hline
& A & B& C& D& E& F & G \\ 
&1-7 & 2-6& 3-5& 4-4& 5-3& 6-2 & 7-1 \\
\hline
jack&   100.0&	100.0&	100.0&	100.0&	100.0&	100.0&	 100.0  \\
pjump&   87.7&	 92.8&	 94.4&	 96.3&	 97.0&	 95.5&	  95.5 \\
skip&    77.4&	 83.7&	 86.8&	 86.8&	 83.7&	 87.5&	  87.5  \\
run&     32.7&	 40.0&	 48.8&	 53.3&	 61.0&	 65.0&	  65.0   \\
walk&    83.6&	 93.3&	 94.4&	 95.0&   96.7&	 96.0&	  96.0   \\
wave1&   92.3&	 99.2&	100.0&  100.0&  100.0&  100.0&   100.0   \\
wave2&   48.4&   65.7&	 75.4&	 76.3&	 84.0&	 86.5&	  86.5	\\
\end{tabular}
\label{accucary_by_actions}}
\end{table*}

\begin{table*}[ht!]
\centering
\caption{Confusion matrix among actions (in \%) according to dataset provided by L. Gorelick et al. for each of the  7 experiments considered (part II).}
\begin{tabular}{c|c| c c c c c c c c}
\hline
& Actions& jack & pjump & skip & run & walk & wave1 & wave2 \\ 
%&1-7 & 2-6& 3-5& 4-4& 5-3& 6-2 & 7-1 \\
\hline
&jack &   \textbf{100}&      0.0&    0.0&    0.0&   0.0&     0.0&     0.0 \\
&pjump&   1.6&     \textbf{87.7}&    0.1&   1.40&   0.0&     8.6&     0.6    \\
&skip &   0.0&      0.7&   \textbf{77.4}&   13.0&   9.3&     0.0&     0.0    \\
A&run &   5.4&      0.7&   48.0&   \textbf{32.7}&  14.0&     0.1&     0.0     \\
1-7&walk &   2.0&      0.0&   10.0&    4.1&  \textbf{83.6}&     0.0&     0.0      \\
&wave1&   1.4&      3.4&    1.0&    0.7&   0.0&    \textbf{92.3}&     1.1         \\
&wave2&   48.0&      0.4&    0.0&    0.1&   0.1&     2.6&    \textbf{48.4}           \\
\hline
&jack &    \textbf{100}&  0.0&     0.0&     0.0&     0.0&     0.0&     0.0      \\
&pjump&     0.3&    \textbf{ 92.8}&     0.0&     2.5&     0.0&     4.3&     0.0   \\
&skip &     0.0&     0.0&     \textbf{83.7}&     9.3&     7.0&     0.0&     0.0    \\
B&run &     1.0&     0.0&     46.8&     \textbf{40.0}&     12.2&     0.0&     0.0 \\
2-6&walk &     0.0&     0.0&     4.2&     2.5&     \textbf{93.3}&     0.0&     0.0    \\
&wave1&     0.0&     0.7&     0.0&     0.0&     0.0&     \textbf{99.}2&     0.2    \\
&wave2&     31.8&     0.0&     0.0&     0.0&     0.0&     2.5&     \textbf{65.7}   \\
\hline
&jack &        \textbf{100}&      0.0&        0.0&        0.0&        0.0&        0.0&        0.0\\ 
&pjump&        0.0&       \textbf{94.4}&        0.0&        1.2&        0.0&        4.4&        0.0\\ 
&skip &        0.0&        0.0&       \textbf{86.8}&        8.4&        4.8&        0.0&        0.0 \\
C&run &        0.0&        0.0&       41.8&      \textbf{ 48.8}&        9.4&        0.0&        0.0\\ 
3-5&walk &        0.0&        0.0&        3.2&        2.4&      \textbf{ 94.4}&        0.0&        0.0 \\
&wave1&        0.0&        0.0&        0.0&        0.0&        0.0&      \textbf{100.0}&        0.0 \\
&wave2&       24.0&        0.0&        0.0&        0.0&        0.0&        0.6&       \textbf{75.4} \\
\hline
&jack &      \textbf{100}&       0.0&        0.0&        0.0&        0.0&        0.0&        0.0 \\ 
&pjump&       0.0&       \textbf{96.3}&        0.0&        0.5&        0.0&        3.3&        0.0 \\ 
&skip &       0.0&        0.0&       \textbf{86.8}&        7.8&        5.5&        0.0&        0.0  \\
D&run &       0.0&        0.0&       38.3&       \textbf{53.3}&        8.5&        0.0&        0.0 \\ 
4-4&walk &       0.0&        0.0&        3.0&        2.0&       \textbf{95.0}&        0.0&        0.0  \\
&wave1&       0.0&        0.0&        0.0&        0.0&        0.0&      \textbf{100.0}&        0.0  \\
&wave2&      23.3&        0.0&        0.0&        0.0&        0.0&        0.5&       \textbf{76.3}  \\
\hline
\end{tabular}
\label{confusion_matrix-1}
\end{table*}

\begin{table*}[ht!]
\centering
\caption{Confusion matrix among actions (in \%) according to dataset provided by L. Gorelick et al. for each of the  7 experiments considered (part II).}
\begin{tabular}{c|c| c c c c c c c c}
\hline
& Actions& jack & pjump & skip & run & walk & wave1 & wave2 \\
\hline
&jack &      \textbf{100}&     0.0&     0.0&     0.0&     0.0&     0.0&     0.0 \\ 
&pjump&        0.0&    \textbf{97.0}&     0.0&     0.7&     0.0&     2.3&     0.0 \\ 
&skip &        0.0&     0.0&    \textbf{83.7}&     9.0&     7.3&     0.0&     0.0  \\
E&run &        0.0&     0.0&    34.7&    \textbf{61.0}&     4.3&     0.0&     0.0 \\ 
5-3&walk &        0.0&     0.0&     0.0&     3.3&    \textbf{96.7}&     0.0&     0.0  \\
&wave1&        0.0&     0.0&     0.0&     0.0&     0.0&    \textbf{100.0}&     0.0  \\
&wave2&       16.0&     0.0&     0.0&     0.0&     0.0&     0.0&    \textbf{84.0}  \\
\hline
&jack &      \textbf{100}&      0.0&      0.0&   0.0&   0.0&      0.0&    0.0     \\ 
&pjump&      0.0&    \textbf{ 95.5}&      0.0&   0.0&   0.0&     4.50&    0.0     \\ 
&skip &      0.0&      0.0&     \textbf{87.5}&   8.0&   4.50&     0.0&    0.0  \\
F&run &      0.0&      0.0&     31.5&  \textbf{65.0}&   3.50&     0.0&    0.0     \\ 
6-2&walk &      0.0&      0.0&      0.0&   4.0&   \textbf{96.0}&     0.0&    0.0     \\
&wave1&      0.0&      0.0&      0.0&   0.0&   0.0&      \textbf{100}&    0.0     \\
&wave2&      13.5&     0.0&      0.0&   0.0&   0.0&      0.0&   \textbf{86.5}&   \\
\hline
&jack &      \textbf{100}&    0.0&   0.0&   0.0&   0.0&    0.0&   0.0   \\ 
&pjump&      0.0&   \textbf{95.5}&   0.0&   0.0&   0.0&    4.5&   0.0   \\ 
&skip &      0.0&    0.0&   \textbf{88.0}&    8.0&   4.5&    0.0&   0.0    \\
G&run &      0.0&    0.0&   32.0&   \textbf{65.0}&   3.5&    0.0&   0.0   \\ 
7-1&walk &      0.0&    0.0&   0.0&   4.0&  \textbf{ 96.0}&   0.0&   0.0    \\
&wave1&      0.0&    0.0&   0.0&   0.0&   0.0&    \textbf{100}&   0.0    \\
&wave2&      14.0&   0.0&   0.0&   0.0&   0.0&    0.0&   \textbf{87.0}   \\
\end{tabular}
\label{confusion_matrix-2}
\end{table*}

Finally, we provide some computations made on
the OU-SIRT-B dataset\footnote{http://www.am.sanken.osaka-u.ac.jp/BiometricDB/GaitTM.html}, where each person has several samples with different combinations of clothes.
%(see Table \ref{abt_clottes} and Table \ref{combination_code}). %
This dataset has  samples from  68 people, taken using a treadmill. It provides the silhouettes, that is, we do not have the videos to apply our own background subtraction algorithm.
Each person may not have samples for all the clothes combinations. Given the number of samples per person (max $27$ and min $9$) it is possible to design several experiments.
As an example, we made a group for similar lower clothes in order to organize the training and the test sets (see Table \ref{cluster}). 

\begin{table*}[ht!]
\centering
\caption{Cluster according to similar lower clothes used.}
\begin{tabular}{l l l l c c}
\hline
 Cluster  (code)  & Lower clothes  &   Upper clothes        \\
\hline
&&&\\
2349ABCXY  &  RP: regular pant  &   2) HS     &      HS: Half Shirt    \\ 
                   &                    &   3) HS,Ht  &      FS: Full Shirt    \\
                   &                    &   4) HS,Cs  &      Ht: Hat           \\    
                   &                    &   9) FS     &      PK: Parka         \\    
                   &                    &   A) PK     &      Cs: Casquette Cap \\
                   &                    &   B) DJ     &      DJ: Down Jacket   \\ 
                   &                    &   C) DJ,Mf  &      Mf: Muffler       \\   
                   &                    &   X) FS,Ht  &           \\            
                   &                    &   Y) FS,Cs  &           \\           
\hline
\end{tabular}
\label{cluster}
\end{table*}

According to clusters in Table \ref{cluster} we took three samples to train (the samples with clothes combination 2, 3 and 4)
and two to test (the samples with clothes combination 9 and A) for each of the first 26 persons. 
Moreover, we took the samples with clothes combination 9, A and B and  to train, and  C and X to test for the rest of the persons, due to people do not have the same combination clothes. Considering only the lower part of the body, we obtain a rate of $79.4$\% of accuracy.

More examples and the source code written in Matlab can be obtained visiting our web page\footnote{http://grupo.us.es/cimagroup/}.

\section{Conclusion}\label{conclusions}

In this paper, we  have presented a
persistent-homology-based 
signature successfully applied in the past to gait recognition. 
We have  shown that  such topological signature can be applied to  any periodic motion (not only gait), such as running or jumping. 
We have  formally proved that such signature is robust to noise and independent to the number of periods used to compute it. 

%In this paper we have presented an algorithm for periodic motion recognition, a technique with special attention in tasks of video surveillance. 
% We have used persistent  homology to model the motion,  similar  as we did in our previous approaches for gait recognition.
% 
%Regarding gait recognition, the topological signature have been tested in this paper using only the lowest fourth part of the body silhouette. Then, the effects of variations unrelated to the gait in the upper body part, which are very frequent in real scenarios, decrease considerably. 
%We have also tested our topological signature to other kind of  motions, showing its applicability  to periodic motions in general, not only gait.
%We have also formally proved that our topological signature is robust to small perturbations in the input data and does not depend on the number of periods  contained in the periodic motion sequence.

\vspace{0.5cm}
\noindent{\bf Funding.}
This work has been partially funded by the Applied Math department and Institute of Mathematics at the University of Seville, Andalusian project FQM-369 and Spanish project MTM2015-67072-P.

\bibliographystyle{unsrt}  
\bibliography{references}  %%% Remove 

\begin{thebibliography}{10}

\bibitem{bb65346}
Shiqi Yu, Daoliang Tan, and Tieniu Tan.
\newblock A framework for evaluating the effect of view angle, clothing and
  carrying condition on gait recognition.
\newblock In {\em Pattern Recognition, 2006. ICPR 2006. 18th International
  Conference on}, volume~4, pages 441--444. IEEE, 2006.

\bibitem{lee2014time}
Chin~Poo Lee, Alan~WC Tan, and Shing~Chiang Tan.
\newblock Time-sliced averaged motion history image for gait recognition.
\newblock {\em Journal of Visual Communication and Image Representation},
  25(5):822--826, 2014.

\bibitem{lamar2012human}
Javier Lamar-Leon, Edel Garcia-Reyes, and Rocio Gonzalez-Diaz.
\newblock Human gait identification using persistent homology.
\newblock In {\em Progress in Pattern Recognition, Image Analysis, Computer
  Vision, and Applications - 17th Iberoamerican Congress, {CIARP} 2012, Buenos
  Aires, Argentina, September 3-6, 2012. Proceedings}, pages 244--251, 2012.

\bibitem{zhang2013score}
Yuanyuan Zhang, Shuming Jiang, Zijiang Yang, Yanqing Zhao, and Tingting Guo.
\newblock A score level fusion framework for gait-based human recognition.
\newblock In {\em Multimedia Signal Processing (MMSP), 2013 IEEE 15th
  International Workshop on}, pages 189--194. IEEE, 2013.

\bibitem{rida2016gait}
Imad Rida, Somaya Almaadeed, and Ahmed Bouridane.
\newblock Gait recognition based on modified phase-only correlation.
\newblock {\em Signal, Image and Video Processing}, 10(3):463--470, 2016.

\bibitem{wu2017comprehensive}
Zifeng Wu, Yongzhen Huang, Liang Wang, Xiaogang Wang, and Tieniu Tan.
\newblock A comprehensive study on cross-view gait based human identification
  with deep cnns.
\newblock {\em IEEE transactions on pattern analysis and machine intelligence},
  39(2):209--226, 2017.

\bibitem{bb68418}
Changhong Chen, Jimin Liang, Heng Zhao, Haihong Hu, and Jie Tian.
\newblock Frame difference energy image for gait recognition with incomplete
  silhouettes.
\newblock {\em Pattern Recognition Letters}, 30(11):977--984, 2009.

\bibitem{leon2013gait}
Javier Lamar-Leon, Andrea Cerri, Edel Garia-Reyes, and Rocio Gonzalez-Diaz.
\newblock Gait-based gender classification using persistent homology.
\newblock In {\em Progress in Pattern Recognition, Image Analysis, Computer
  Vision, and Applications - 18th Iberoamerican Congress, {CIARP} 2013, Havana,
  Cuba, November 20-23, 2013, Proceedings, Part {II}}, pages 366--373, 2013.

\bibitem{ciarp2014}
Javier Lamar-Leon, Raul Alonso-Baryolo, Edel Garcia-Reyes, and Rocio
  Gonzalez-Diaz.
\newblock Gait-based carried object detection using persistent homology.
\newblock In {\em Progress in Pattern Recognition, Image Analysis, Computer
  Vision, and Applications - 19th Iberoamerican Congress, {CIARP} 2014, Puerto
  Vallarta, Mexico, November 2-5, 2014. Proceedings}, pages 836--843, 2014.

\bibitem{icpr2014}
Javier Lamar-Leon, Raul Alonso-Baryolo, Edel Garcia-Reyes, and Roio
  Gonzalez-Diaz.
\newblock Topological features for monitoring human activities at distance.
\newblock In {\em Activity Monitoring by Multiple Distributed Sensing - Second
  International Workshop, {AMMDS} 2014, Stockholm, Sweden, August 24, 2014},
  pages 40--51, 2014.

\bibitem{icpr2016}
Javier Lamar-Leon, Raul~Alonso Baryolo, Edel Garcia-Reyes, and Rocio
  Gonzalez-Diaz.
\newblock Persistent homology-based gait recognition robust to upper body
  variations.
\newblock In {\em 23rd International Conference on Pattern Recognition, {ICPR}
  2016, Canc{\'{u}}n, Mexico, December 4-8, 2016}, pages 1083--1088, 2016.

\bibitem{bashir2010gait}
Khalid Bashir, Tao Xiang, and Shaogang Gong.
\newblock Gait recognition without subject cooperation.
\newblock {\em Pattern Recognition Letters}, 31(13):2052--2060, 2010.

\bibitem{yonghzhen}
Zhen Zhou, Yongzhen Huang, Liang Wang, and Tieniu Tan.
\newblock Exploring generalized shape analysis by topological representations.
\newblock {\em Pattern Recognition Letters}, 87:177--185, 2017.

\bibitem{MJ}
Maria~Jose Jimenez, Bel{\'{e}}n Medrano, David~S. Monaghan, and Noel~E.
  O'Connor.
\newblock Designing a topological algorithm for 3d activity recognition.
\newblock In {\em Computational Topology in Image Context - 6th International
  Workshop, {CTIC} 2016, Marseille, France, June 15-17, 2016, Proceedings},
  pages 193--203, 2016.

\bibitem{RELATED2016}
V.~{Venkataraman}, K.~N. {Ramamurthy}, and P.~{Turaga}.
\newblock Persistent homology of attractors for action recognition.
\newblock In {\em 2016 IEEE International Conference on Image Processing
  (ICIP)}, pages 4150--4154, Sep. 2016.

\bibitem{related2016-1}
A.~{Dirafzoon}, N.~{Lokare}, and E.~{Lobaton}.
\newblock Action classification from motion capture data using topological data
  analysis.
\newblock In {\em 2016 IEEE Global Conference on Signal and Information
  Processing (GlobalSIP)}, pages 1260--1264, Dec 2016.

\bibitem{related2015}
Mikael Vejdemo-Johansson, Florian~T. Pokorny, Primoz Skraba, and Danica Kragic.
\newblock Cohomological learning of periodic motion.
\newblock {\em Applicable Algebra in Engineering, Communication and Computing},
  26(1):5--26, Mar 2015.

\bibitem{compTopologyBook}
Herbert Edelsbrunner and John Harer.
\newblock {\em Computational topology: an introduction}.
\newblock American Mathematical Soc., 2010.

\bibitem{ghrist}
Robert Ghrist.
\newblock Barcodes: The persistent topology of data.
\newblock {\em Bull. Amer. Math. Soc. 45 (2008), 61-75}, 45:61--75, 2008.

\bibitem{chazal}
Fr{\'e}d{\'e}ric Chazal, Vin de~Silva, and Steve Oudot.
\newblock Persistence stability for geometric complexes.
\newblock {\em Geometriae Dedicata}, 173(1):193--214, 2014.

\bibitem{mukres}
James~R Munkres.
\newblock {\em Elements of algebraic topology}, volume~2.
\newblock Addison-Wesley Reading, 1984.

\bibitem{singh2009biometric}
Shamsher Singh and KK~Biswas.
\newblock Biometric gait recognition with carrying and clothing variants.
\newblock In {\em Pattern Recognition and Machine Intelligence}, pages
  446--451. Springer, 2009.

\bibitem{lishani2014haralick}
Ait~O Lishani, Larbi Boubchir, and Ahmed Bouridane.
\newblock Haralick features for gei-based human gait recognition.
\newblock In {\em Microelectronics (ICM), 2014 26th International Conference
  on}, pages 36--39. IEEE, 2014.

\end{thebibliography}
\end{document}